\documentclass{article}

\usepackage{microtype}
\usepackage{graphicx}
\usepackage{subcaption}
\usepackage{booktabs} 

\usepackage{hyperref}

\usepackage[accepted]{icml2026}

\usepackage{mathtools}
\usepackage{amsmath, amssymb, amsthm, bbm, amsfonts}
\usepackage{physics}
\usepackage{hyperref}       \usepackage{caption}
\usepackage{tensor}
\usepackage[most]{tcolorbox}
\usepackage{empheq}
\usepackage{enumitem}
\setlist{nosep, noitemsep, leftmargin=0.75em}
\usepackage{float}
\usepackage{nicefrac}       \usepackage{microtype}      \usepackage{tikz}
\usepackage{xfrac}

\hypersetup{colorlinks, breaklinks=true, urlcolor=blue, linkcolor=blue, citecolor=blue}

\DeclareMathOperator{\sign}{sign}

\DeclareMathOperator*{\argmin}{arg\,min}

\newcommand{\RR}{\ensuremath{\mathbb{R}}}

\newcommand{\s}{\ensuremath{\boldsymbol{s}}}

\renewcommand{\eqref}[1]{\textup{{\normalfont Eq.~(\ref{#1}}\normalfont)}}

\newcommand{\qlatent}{\ensuremath{\tau_{\ell}}}

\newcommand{\qstar}{{\ensuremath{q^\star}}}

\newcommand{\vv}{{\ensuremath{\boldsymbol{v}}}}

\newcommand{\z}{\ensuremath{\boldsymbol{z}}}
\newcommand{\e}{\ensuremath{\boldsymbol{e}}}
\newcommand{\x}{\ensuremath{\boldsymbol{x}}}
\newcommand{\w}{\ensuremath{\boldsymbol{w}}}
\newcommand{\btheta}{\ensuremath{\boldsymbol{\theta}}}
\newcommand{\wstar}{\ensuremath{\boldsymbol{w}_\star}}

\newcommand{\bxi}{\ensuremath{\boldsymbol{\xi}}}

\renewcommand{\tilde}{\widetilde}

\newcommand{\advtrainingcost}{\ensuremath{\varepsilon_t}}

\newcommand{\adverrproper}{\ensuremath{E_{\mathrm{rob}}^{\mathrm{cns}}}}
\newcommand{\bounderrproper}{\ensuremath{E_{\mathrm{bnd}}^{\mathrm{cns}}}}

\newcommand{\advgenerr}{\ensuremath{E_{\mathrm{rob}}}}

\newcommand{\what}{\ensuremath{\hat{\boldsymbol{w}}}}
\newcommand{\thetahat}{\ensuremath{\hat{\boldsymbol{\theta}}}}

\newcommand{\y}{\ensuremath{\boldsymbol{y}}}

\newcommand{\uu}{\ensuremath{\boldsymbol{u}}}

\newcommand{\gaussvecone}{\ensuremath{\boldsymbol{g}}}
\newcommand{\gaussvectwo}{\ensuremath{\boldsymbol{h}}}

\newcommand{\dataidx}{i}

\newcommand{\nsamples}{n}

\newcommand{\lossfun}{\ell}
\newcommand{\regfun}{r}

\newcommand{\defnorm}{s}
\newcommand{\attnorm}{q}
\newcommand{\defend}{\boldsymbol{v}}
\newcommand{\attack}{\boldsymbol{\delta}}
\newcommand{\revision}[1]{#1}

\newcommand{\idmat}{\ensuremath{\operatorname{Id}}}
\newcommand{\featmat}{\ensuremath{\operatorname{F}}}

\usepackage[capitalize]{cleveref}

\crefname{assumption}{Assumption}{Assumptions}

\theoremstyle{plain}
\newtheorem{theorem}{Theorem}[section]
\newtheorem{proposition}[theorem]{Proposition}
\newtheorem{lemma}[theorem]{Lemma}
\newtheorem{corollary}[theorem]{Corollary}
\theoremstyle{definition}
\newtheorem{definition}[theorem]{Definition}
\newtheorem{assumption}[theorem]{Assumption}
\theoremstyle{remark}
\newtheorem{remark}[theorem]{Remark}

\usepackage[disable,textsize=tiny]{todonotes}

\icmltitlerunning{On the Existence of Consistent Adversarial Attacks in High-Dimensional Linear Classification}

\begin{document}

\twocolumn[
  \icmltitle{On the Existence of Consistent Adversarial Attacks\protect\\
  in High-Dimensional Linear Classification}

\icmlsetsymbol{equal}{*}

  \begin{icmlauthorlist}
    \icmlauthor{Matteo Vilucchio}{idephicsepfl}
    \icmlauthor{Lenka Zdeborová}{spocepfl}
    \icmlauthor{Bruno Loureiro}{ens}
  \end{icmlauthorlist}

  \icmlaffiliation{idephicsepfl}{Information, Learning \& Physics Laboratory, École Polytechnique Fédérale de Lausanne (EPFL), Lausanne, Switzerland}
  \icmlaffiliation{spocepfl}{Statistical Physics of Computation Laboratory, École Polytechnique Fédérale de Lausanne (EPFL), Lausanne, Switzerland}
  \icmlaffiliation{ens}{Département d'Informatique de l'École Normale Supérieure (DI ENS), CNRS, PSL University, Paris, France}

  \icmlcorrespondingauthor{Matteo Vilucchio}{matteo.vilucchio@epfl.ch}

\icmlkeywords{Machine Learning, ICML, Adversarial Robustness, Consistent Attacks, High-dimensional Statistics}

  \vskip 0.3in
]

\printAffiliationsAndNotice{}  

\begin{abstract}
What fundamentally distinguishes an adversarial attack from a misclassification due to limited model expressivity or finite data?
In this work, we investigate this question in the setting of high-dimensional binary classification, where statistical effects due to limited data availability play a central role.
We introduce a new error metric that precisely captures this distinction, quantifying model vulnerability to consistent adversarial attacks --- perturbations that preserve the ground-truth labels.
Our main technical contribution is an exact and rigorous asymptotic characterization of these metrics in both well-specified models and latent space models, revealing different vulnerability patterns compared to standard robust error measures.
The theoretical results demonstrate that overparameterization affects vulnerability to label-preserving perturbations in a metric- and regime-dependent manner, offering theoretical insight into the mechanisms underlying model sensitivity to adversarial attacks.
\end{abstract}

\section{Introduction}
\label{sec:introduction}
Machine learning models, despite their remarkable performance across various domains, remain vulnerable to adversarial examples --- inputs specifically crafted to mislead models while appearing innocuous to humans. While adversarial robustness has attracted significant research attention, a critical distinction often overlooked is between \textit{consistent} (or \emph{proper}) and \textit{inconsistent} (or \emph{improper}) adversarial examples. Consistent adversarial examples maintain the ground-truth label despite perturbations, whereas inconsistent ones change the true classification.
\begin{figure}
\centering
\includegraphics[width=0.75\linewidth]{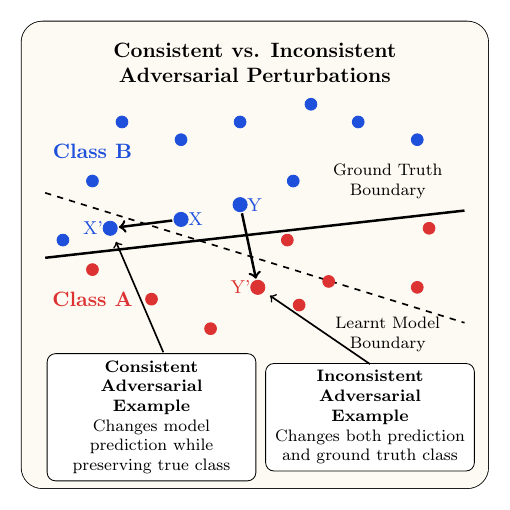}
\caption{Illustration of the difference between consistent Adversarial Perturbation and inconsistent Adversarial Perturbation.}
\label{fig:cartoon}
\end{figure}
Consider the classic panda example from \citet{GSS15}: an image that, after subtle perturbations, is misclassified by a neural network but still depicts a panda to humans (consistent attack). If perturbations transformed the image to genuinely resemble a different animal, it would be inconsistent. This distinction is crucial: vulnerability to consistent attacks represents genuine model failure to capture invariant features.

The assumption that adversarial perturbations do not alter the true class (\textit{i.e.} remain consistent) underlies most practical approaches to adversarial robustness in computer vision \citep{Mad+18}.
While this assumption has been explored in theoretical works on robust generalization \citep{Donhauser2021,raghunathan2020understanding}, a mathematical understanding of such attacks, including their existence and effectiveness in tricking even simple linear classifiers, remains elusive.

Following a large body of work in high-dimensional statistics \citep{krogh1991simple, SST92, bean2013optimal, thrampoulidis2018precise, aubin_generalization_2020, vilucchio2024asymptotic}, we analyze this problem through the lens of exact asymptotics of linear classifiers.
We develop novel metrics that precisely quantify vulnerability to both consistent and inconsistent adversarial attacks.
We define and analyze these metrics in two complementary settings: first, a \textit{well-specified} model where all input covariates are directly available; second, a \textit{latent space} model where the available covariates are feature transformations of underlying latent variables.
For both settings, we derive closed-form expressions for the consistent robustness metrics in the high-dimensional limit—where the latent space dimension $d$, the number of features $p$, and the sample size $n$ all scale to infinity at fixed ratios.
In the latent space model, we further derive exact asymptotic descriptions for the performance of robust empirical risk minimization \cite{GSS15,Mad+18},
the most widely adopted method for finding a robust model.

Furthermore, the effect of over or under-parameterization --- using more or less parameters than strictly necessary to encode the data features --- is still unclear in the adversarial settings.
\citet{ribeiro2023overparameterized} consider the regression case for squared loss, but in classification the question is still open.
While overparameterization generally improves standard generalization \citep{goldt_gaussian_2021,hastie2022surprises}, its effects on adversarial robustness are less understood, particularly when considering consistent versus inconsistent attacks.
Conventional wisdom suggests that overparameterized models might be more vulnerable to adversarial examples due to their flexibility in fitting noise.
However, our analysis reveals a more nuanced picture: the effect of overparameterization is both metric- and regime-dependent, and the finite-range trends of consistent robust and boundary errors can turn over in the extreme overparameterized limit.
See~\Cref{sec:app:further-rel-works} for a further discussion of related works.

Our \textbf{main contributions} can be summarized as\footnote{Code reproducing all results is available at
\mbox{\url{https://github.com/IdePHICS/adv-consistent}}.}
\begin{enumerate}
    \item We establish necessary and sufficient conditions for the existence of consistent perturbations in two classes of binary classifiers: well-specified linear classifiers, and a latent variable model that accounts for misspecification and overparametrization in linear estimation, independently of the data distribution. Under a Gaussian design, this leads to an exact formula for the probability that consistent attacks exist in these models.

    \item We introduce novel consistent robust error metrics quantifying the impact of consistent attacks. For the classifiers of interest, we derive asymptotic formulas that exactly characterize their high-dimensional limits under a Gaussian design assumption.

    \item We study how robust empirical risk minimization can mitigate consistent attacks in this high-dimensional limit, for both the well-specified and latent variable model. For the latter, this requires a novel exact asymptotic characterization of the robust ERM estimator under misspecification, of independent interest.
\end{enumerate}
Our work reveals that overparameterization plays a nuanced but crucial role in building resistance against consistent adversarial attacks.
Contrary to conventional wisdom, our theoretical analysis demonstrates that higher degrees of parameterization can be beneficial for overall robustness, though this benefit must be balanced against increased vulnerability on specific subsets of inputs.
These insights can provide guidance for system design, highlighting the importance of considering the consistent/inconsistent attack distinction when evaluating and optimizing model robustness.

\paragraph{Notations}
We denote vectors by bold letters $\x\in\mathbb{R}^{d}$.
$\mathbb{S}^{d-1}(r)=\{\x\in\mathbb{R}^{d}: ||\x||_{2}=r\}$ denotes the Euclidean sphere of radius $r$, and ${\rm span}(\x) = \x\mathbb{R} = \{\mu \x,\mu\in\mathbb{R}\}$.
For $q\geq 1$, $||\x||_{q}=(\sum_{j=1}^{d}|x_{j}|^{q})^{\sfrac{1}{q}}$ denotes the $\ell^{q}$-norm, and $B_{q}(r)=\{\x\in\mathbb{R}^{d}:||\x||_{q}\leq r\}$ the $\ell^{q}$-ball of radius $r>0$.
We denote by $\attnorm^{\star}$ the dual of $q$ in the $\ell^{\attnorm}$ sense: the number $\attnorm^\star$ such that $\sfrac{1}{\attnorm}+\sfrac{1}{\attnorm^{\star}}=1$. We denote by $\mathcal{N}(0,1)$ the standard normal distribution, and $\mathbb{P}[Z\leq t]=\Phi(t)$ its c.d.f.
Additionally $\mathcal{P}_{f(\cdot)}$ represents the proximal operator of the function $f:\mathbb{R} \to \mathbb{R}$.

\section{Consistent Adversarial Perturbations and How To Quantify Them}
\label{sec:existence-proper-adversarial}
As motivated in \Cref{sec:introduction}, the key distinction between an adversarial attack and a random perturbation of the data is the underlying assumption that adversarial attacks leave the ground truth data distribution unchanged.
We formalize this notion in the context of binary classifiers.

\revision{Consider a binary classification task defined by a covariate distribution $\mu$ over $\mathbb{R}^{d}$, a target score $g_{\star}:\mathbb{R}^{d}\to\mathbb{R}$, and a link $\varphi_{\star}$ such that $f_{\star}=\varphi_{\star}\circ g_{\star}$ and $f_{\star}(\x)=\mathbb{P}(y=+1|\x)$.}

\begin{definition}[Consistent attack]
\label{def:consistent}
Let $f_{\star}=\varphi_{\star}\circ g_{\star}$ and $\hat{f}:\mathbb{R}^{d}\to [0,1]$ denote two binary classifiers, referred to as the \emph{target} and the \emph{model},
$\x\in\mathbb{R}^{d}$ a covariate and $\hat{y} : [0,1] \to \{\pm 1\}$ a decision rule associated to $\hat{f}$.
We say a perturbation $\attack\in B_{\attnorm}(\varepsilon)$ of the model $\hat{f}$ is a \emph{consistent} adversarial attack with respect to the target $f_{\star}$,
the covariate $\x\in\mathbb{R}^{d}$ and the decision rule $\hat{y}$ if the following conditions hold:
\begin{itemize}
    \item \textbf{Model deception:} $\hat{y}(\hat{f}(\x)) \neq \hat{y}(\hat{f}(\x+\attack))$.
    \item \textbf{Target-score invariance:} $g_{\star}(\x) = g_{\star}(\x+\attack)$.
\end{itemize}
If target-score invariance fails the attack is \emph{inconsistent}.
\end{definition}
See \Cref{fig:cartoon} for an illustration of a consistent vs. inconsistent attack in the case of linear classifiers.
{
\begin{remark}\label{rmk:consistent-prob}
For deterministic classifiers, target-score invariance implies target-label invariance, but not conversely:
{\setlength{\abovedisplayskip}{2pt}\setlength{\belowdisplayskip}{2pt}
\begin{equation*}\small
\{\attack:g_{\star}(\x)=g_{\star}(\x+\attack)\}
\subseteq \{\attack:\hat y(f_{\star}(\x))=\hat y(f_{\star}(\x+\attack))\}.
\end{equation*}}
For probabilistic classifiers, it instead preserves the conditional label distribution $\mathbb{P}(y|\x)$, not individual random label realizations---the natural statistical analogue. Hence our consistent metrics lower-bound label-preserving vulnerability, while the standard robust error upper-bounds it.
\end{remark}
}
We introduce metrics to quantify these properties.
\begin{definition}[Adversarial errors]
\label{def:metrics}
Let $(\mu,g_{\star},\varphi_{\star})$ denote a binary classification task. Given a classifier $\hat{f}:\mathbb{R}^{d}\to[0,1]$ and its associated predictor $\hat{y}(\x)=\sign(2\hat f(\x)-1)$, we define the following three metrics
\begin{itemize}
\item \textbf{Robust error}: This is the standard notion of robust generalization error in the adversarial literature \cite{clarysse2022adversarial,dohmatob2024consistent,dohmatob2023robust}, and simply quantifies how vulnerable $\hat{f}$ is to arbitrary perturbations in a $\ell^{\attnorm}-$ball:
{\small
\begin{equation}\label{eq:classical-robust-error}
\advgenerr
= \mathbb{E}_{(\x,y)}\qty[
        \max_{\attack\in B_{\attnorm}(\varepsilon)}
        \mathbbm{1}\qty{
            y
            \neq
            \hat{y}(\x + \attack)
        }
    ] \,,
\end{equation}
}
\item \textbf{Consistent robust error:} The standard robust error considers both consistent and inconsistent perturbations. To quantify just the role of consistent attacks, we define the \emph{consistent robust error} by excluding inconsistent perturbations:
{\begin{equation}\label{eq:def-misclassif-proper}
    \adverrproper
=
    \mathbb{E}_{(\x,y)}\qty[
        \max_{
            \substack{
                \attack\in B_{\attnorm}(\varepsilon): \\
                g_{\star}(\x)=g_{\star}(\x+\attack)
            }
        } \!\!\!
        \mathbbm{1}\qty{
            y
            \neq
            \hat{y}(\x + \attack)
        }
    ] \,,
\end{equation}
}
Note that the critical difference between $\advgenerr$ and $\adverrproper$ is the target-score invariance constraint from~\Cref{def:consistent} (cf.~\Cref{rmk:consistent-prob}).
\item \textbf{Consistent boundary error:} Finally, note that the consistent robust error does not distinguish between labels that are originally misclassified by the model and labels that become misclassified under the attack perturbation. This motivates the introduction of a more nuanced metric, accounting only for labels that are misclassified due to the attack:
\begin{equation}\label{eq:boundary-proper}
    \!\!\!\!\bounderrproper
\!= \!
    \mathbb{E}_{(\x,y)}\qty[
        \max_{
            \substack{
                \attack\in B_{\attnorm}(\varepsilon): \\
                g_{\star}(\x)=g_{\star}(\x+\attack)
            }
        } \!\!\!\!\!
\mathbbm{1}\qty{
        \genfrac{}{}{0pt}{}{
            y
            \neq
            \hat{y}(\x + \attack)
        }{
            y = \hat{y}(\x)
        }
        }
] .
\end{equation}
\end{itemize}
\end{definition}
\begin{remark}
\label{rmk:bounds}
    For each $(\x,y)$, the feasible set of consistent perturbations is a subset of $B_{\attnorm}(\varepsilon)$, while the boundary-error integrand additionally requires $y=\hat y(\x)$. The corresponding maxima are therefore ordered pointwise, and hence $0\leq \bounderrproper \leq \adverrproper \leq \advgenerr$.
\end{remark}
\section{Consistent Attacks In Well-Specified Linear Classification}
\label{sec:wellspe}
Despite an established literature studying robust training schemes, the fundamental properties of consistent attacks remain poorly understood. Our goal in the following is to fill this gap by studying their behavior in the context of high-dimensional binary linear classifiers.
\begin{definition}[Linear classifiers] A linear binary classifier in $\mathbb{R}^{d}$ is a function
\begin{align}
\label{eq:probit-def}
    f_{\w}(\x) = \mathbb{P}_{\w}(y=+1|\x) = \varphi(g_{\w}(\x)) \,,
\end{align}
defined by the \emph{weight vector} $\w\in\mathbb{R}^{d}$, the score $g_{\w}(\x)=\langle \w,\x\rangle$, and a monotonic \emph{link function} $\varphi:\mathbb{R}\to [0,1]$.
\end{definition}
The class of linear binary classifiers encompasses several models of interest in statistics, such as the logit $\varphi(t)=(1+e^{-t})^{-1}$, the probit $\varphi(t)=\sfrac{1}{2}(\erf(t)+1)$ and the noiseless $\varphi(t)=\mathbbm{1}\qty{t\geq 0}$ model.

\subsection{Geometry of Consistent Attacks}
\label{sec:geometry:wellsp}
As a first step, we consider the geometry of consistent attacks in the class of linear classifiers. Let $f_{\wstar}$ denote a reference linear classifier with weights $\wstar\in\mathbb{S}^{d-1}(\sqrt{d})$ and link function $\varphi_{\star}$, which we
refer to as the \emph{ground-truth}.
For the linear target score $g_{\wstar}(\x)=\langle\wstar,\x\rangle$, target-score invariance is equivalent to $\langle \attack,\wstar\rangle = 0$, \textit{i.e.} the attack must be orthogonal to
$\wstar$.
Therefore, the set of admissible consistent adversarial attacks with respect to the target classifier defines a hyperplane:
\begin{align}
\label{eq:admissible}
    H_{\attnorm}(\varepsilon) \coloneqq \{\attack\in B_{\attnorm}(\varepsilon): \langle \wstar,\attack\rangle =0\} \,.
\end{align}
Consider a second linear classifier $f_{\hat{\w}}$ with weights $\hat{\w}\in\mathbb{R}^{d}$ and link function $\varphi$, which we
refer to as the \emph{model}. A successful attack should flip the predictor labels $\hat{y}(\x){\neq}\hat{y}(\x+\attack)$. For the standard decision function $\hat{y}(\x){=}\sign(2f_{\hat{\w}}(\x)-1)$ this
is equivalent to
$\langle \hat{\w},\x\rangle (\langle \hat{\w},\x\rangle + \langle \hat{\w},\attack\rangle)\leq 0$. This is the case if and only if:
\begin{align}
\label{eq:modelflip}
{
\!\!|\langle \hat{\w},\attack\rangle| \!>\! |\langle \hat{\w},\x\rangle| , \:
\sign(\langle \hat{\w},\attack\rangle) \!=\! -\sign(\langle \hat{\w},\x\rangle) .
}
\end{align}
In other words,
to flip the model prediction, an attacker must have an anti-parallel component to the predictor weights and exceed the prediction margin $|\langle \hat{\w},\x\rangle|$.
Combining these observations, we can derive the following geometrical characterization for the existence of consistent perturbations.
\begin{proposition}[Existence of consistent attack]
\label{prop:exis}
\begin{figure}[t]
\centering
\begin{tikzpicture}[scale=1.3,>=latex,line join=round,font=\scriptsize]
\draw[gray,dashed,thick] (-1.35,0) -- (1.35,0)
        node[right,black,inner sep=1pt] {$\langle\wstar,\x\rangle{=}0$};
\draw[gray!50!red,dashed,thick] (-1.35,1.0) -- (1.35,1.0);
\draw[red!70!black,thick] (-0.65,1.08) -- (-0.65,0.92);
  \draw[red!70!black,thick] ( 1.25,1.08) -- ( 1.25,0.92);
  \node[red!70!black,below,inner sep=1pt] at (1.25,0.92) {$H_{\attnorm}(\varepsilon)$};
\draw[thick] (-0.54,1.2555) -- (0.4,-0.93);
  \node[above right,inner sep=1pt,black] at (-0.75,-0.95) {$\langle\what,\x\rangle{=}0$};
\draw[->,thick] (0,0) -- (0,0.85)   node[left,inner sep=1pt] {$\wstar$};
  \draw[->,thick] (0,0) -- (0.95,0.4) node[below right,inner sep=0pt] {$\what$};
\draw[->,red,very thick] (0.3,1.0) -- (-0.55,1.0)
        node[midway,above,inner sep=1pt] {$\attack$};
\fill[black] (0.3,1.0) circle (1.1pt) node[above right,inner sep=1pt] {$\x$};
\end{tikzpicture}
\caption{
    Geometry of consistent attacks (\Cref{sec:geometry:wellsp}). Target-score invariance forces the perturbation to lie in the hyperplane $H_{\attnorm}(\varepsilon)=\{\attack\in B_{\attnorm}(\varepsilon):\langle\wstar,\attack\rangle=0\}$ orthogonal to the target weights $\wstar$ (here a segment, delimited by the two red ticks), so $\attack$ stays parallel to the target boundary (upper dashed line). A consistent attack succeeds when it pushes $\x$ across the model boundary $\langle\what,\x\rangle=0$ without leaving $H_{\attnorm}(\varepsilon)$.
}
\label{fig:geometry-consistent}
\end{figure}

Consider two linear classifiers defined by the weights $\wstar\in\mathbb{S}^{d-1}(\sqrt{d})$ and $\hat{\w}\in\mathbb{R}^{d}$.
Let $\x\in\mathbb{R}^{d}$ denote a covariate,
and assume $\langle \hat{\w},\x\rangle \neq 0$.
Then, a consistent attack $\attack\in B_{\attnorm}(\varepsilon)$ with respect to $\wstar,\x\in\mathbb{R}^{d}$ and the decision function $\hat{y}(\x)=\sign(2f_{\hat{\w}}(\x)-1)$ exists if and only if:
\begin{align}
\label{eq:cond-well-spec}
      \varepsilon~d^{\star}_{q^{\star}}(\hat{\w}_{\perp}) \geq |\langle \hat{\w},\x\rangle | \,,
\end{align}
where
$\hat{\w}_{\perp} = \hat{\w}-\sfrac{1}{d}\langle \wstar,\hat{\w}\rangle\, \wstar$ is the predictor components orthogonal to the target weights,
$d^{\star}_{q^{\star}}(\hat{\w}_{\perp})=\inf_{\mu\in\mathbb{R}}||\hat{\w}_{\perp}-\mu\wstar||_{q^{\star}}$
is the $\ell^{q^{\star}}$ distance to the ${\rm span(\wstar)}$ and $q^{\star}$ is the dual of $q$.
\end{proposition}

\begin{remark}
    For $q=2$, the infimum in the definition of $d^{\star}_{q^{\star}}$ is achieved at $\mu=0$:
    \begin{align}
        d^{\star}_{2}(\hat{\w}_{\perp})= \inf_{\mu\in\mathbb{R}}||\hat{\w}_{\perp}-\mu\wstar||_{2} = ||\hat{\w}_{\perp}||_{2}.
    \end{align}
    \revision{While $d^{\star}_{q^{\star}}(\hat{\w}_{\perp}) \!\leq\! ||\hat{\w}_{\perp}||_{q^{\star}}$ is always an upper bound, it is not tight for $q\!\neq\! 2$, except for particular choices of $\wstar\!\in\!\mathbb{S}^{d-1}(\sqrt{d})$, for instance $\wstar \!=\! \sqrt{d} \e_{1}$. This highlights how the existence of consistent attacks
depends on an interplay between the Euclidean geometry of the constraint set and the $\ell^{q}$ geometry of the adversarial attack.}
\end{remark}
A similar condition to \cref{eq:cond-well-spec} holds for an inconsistent attack, but without the
constraint $\langle \wstar, \attack\rangle =0$. \revision{The direct comparison is}
\begin{align}
\label{eq:bds}
    \revision{d^{\star}_{q^{\star}}(\hat{\w}_{\perp})
    =\inf_{\nu\in\mathbb{R}}\norm{\hat{\w}-\nu\wstar}_{q^{\star}}
    \leq \norm{\hat{\w}}_{q^{\star}}}\,,
\end{align}
this loosens the existence condition, as expected. In particular, the stronger the overlap between the ground-truth and the model $\langle \hat{\w},\wstar\rangle$, the stronger the attack needs to be in order to flip the model prediction, in contrast to inconsistent perturbations which are independent of the ground-truth weights $\wstar$.
This leads to the following corollary.
\begin{corollary}[Existence of inconsistent attack]
\label{cor:incon}
Under the same setting as~\Cref{prop:exis}, an inconsistent adversarial attack exists if and only if:
\begin{align}
\label{eq:cor1}
    \epsilon ||\hat{\w}||_{q^{\star}}\geq |\langle\hat{\w},x\rangle| \,.
\end{align}
\revision{Moreover, every consistent perturbation is feasible for the unrestricted
adversary, and hence $\adverrproper \leq \advgenerr$.}
\end{corollary}
\revision{\Cref{prop:exis} allows us to identify the region in $\mathbb{R}^{d}$ that is vulnerable to consistent attacks.} Indeed, defining the ground-truth orthogonal margin
\begin{align}
\label{eq:def:marginq}
    \kappa_{\attnorm}(\x) = \frac{|\langle \hat{\w},\x\rangle|}{d^{\star}_{q^{\star}}(\hat{\w}_{\perp})} \,,
\end{align}
a covariate $\x\in\mathbb{R}^{d}$ is vulnerable to a consistent $\attack\in H_{\attnorm}(\varepsilon)$ attack if and only if $\varepsilon > \kappa_{\attnorm}(\x)$, and the \emph{vulnerable region} is given by  $\{\x\in \mathbb{R}^{d}: \kappa_{\attnorm}(\x)<\varepsilon\}\subset\mathbb{R}^{d}$ --- a tube around the decision hyperplane.
Note that this reveals two strategies for mitigating consistent adversarial attacks by increasing $\kappa_{\attnorm}$: (a) aligning with the target weights; (b) reducing $d^{\star}_{q^{\star}}(\hat{\w}_{\perp})$.
\revision{While the first option is typically beyond the statistician's control, the second can be achieved by explicitly regularizing training with the norm dual to the attack, which upper-bounds $d^{\star}_{q^{\star}}(\hat{\w}_{\perp})$, see \cref{eq:bds}. This is consistent with previous theoretical results suggesting the use of the dual norm in ERM \citep{YRB19, AFM20, Tsi+24, vilucchio2024geometry}, and is the subject of~\Cref{sec:highd}.}

Another important factor in the margin $\kappa_{\attnorm}(\varepsilon)$ is the interplay between the underlying Euclidean ($\ell^{2}$) geometry defining the classifier and the $\ell^{\attnorm}$-geometry of the adversarial attack. This interplay is best illustrated in the Gaussian case.

\begin{corollary}[\revision{Existence for Gaussian Covariates with General Covariance}]
\label{corr:well:Gauss:general}
In the same setting as~\Cref{prop:exis} with Gaussian covariates $\x \sim \mathcal{N}(\mathbf{0}, \boldsymbol{\Sigma})$, the probability that a consistent attack exists is:
\begin{align}
\label{eq:probGauss-well-spec-general}
    \mathbb{P}\left[\exists\text{ consistent } \attack\in \!H_{\attnorm}(\varepsilon)\right] \!=\! 2\Phi\!\left(\frac{\varepsilon d^{\star}_{q^{\star}}(\hat{\w}_{\perp})}{\sqrt{\hat{\w}^{\top}\boldsymbol{\Sigma}\hat{\w}}}\right)\!-\!1 \,,
\end{align}
where $\Phi$ is the standard normal c.d.f.
\revision{For the noiseless link and nondegenerate target and model fields, let
$\rho=d^{\star}_{q^{\star}}(\hat{\w}_{\perp})/\|\hat{\w}\|_{q^{\star}}$.
The finite-dimensional errors additionally satisfy
$\rho\advgenerr(\varepsilon)\leq\adverrproper(\varepsilon)
=\advgenerr(\rho\varepsilon)\leq\advgenerr(\varepsilon)$.}
\end{corollary}

\begin{figure}
    \centering
    \includegraphics[width=0.345\linewidth]{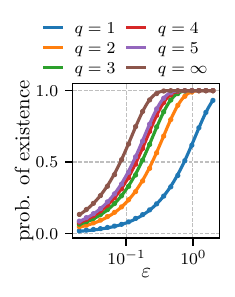}
    \hspace{-1.01em}
    \includegraphics[width=0.345\linewidth]{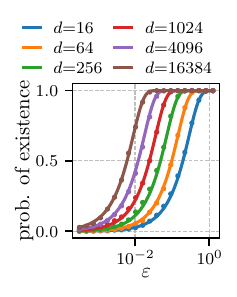}
    \hspace{-1.01em}
    \includegraphics[width=0.345\linewidth]{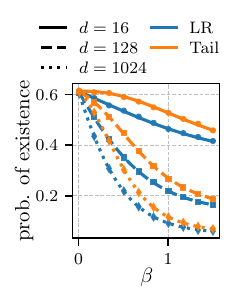}
    \caption{Probability of existence of consistent adversarial attacks for Gaussian covariates via \cref{eq:probGauss-well-spec-general}, with $\wstar=\sqrt{d}\e_{1}$ and $\hat{\w}\in\mathbb{S}^{d-1}(\sqrt{d})$ having correlation $m=0.5$ with $\wstar$.
    \textit{(Left)} Isotropic case $\boldsymbol{\Sigma}\!=\!\sfrac{1}{d}\idmat_{d}$: probability vs $\varepsilon$ for different attack geometries $q$ with $d=10$.
    \textit{(Center)} Dimensional scaling with $q=2$, showing probability vs $\varepsilon$ for varying $d$ for $\boldsymbol{\Sigma}\!=\!\sfrac{1}{d}\idmat_{d}$.
    \textit{(Right)} Power law covariance $\lambda_{i}\sim i^{-\beta}$ for several $d$: probability vs $\beta$ at fixed $\varepsilon=1$ for two predictor types (LR uses top $k{=}10$ modes; Tail uses bottom modes $(k{+}2){:}d$).
    Solid curves show theory; dots show empirical frequencies from $n=10^{4}$ samples satisfying \cref{eq:cond-well-spec}.
    }
    \label{fig:direct-space-prob}
\end{figure}

\begin{remark}
\label{rmk:Gaussian}
The probability in \cref{eq:probGauss-well-spec-general} is non-decreasing in $q$ (non-increasing in $q^{\star}$) and $d$.
To build intuition, we first consider the \emph{isotropic case} $\boldsymbol{\Sigma} \!=\! \sfrac{1}{d}\idmat_{d}$ with a random predictor $\hat{\w} = m\wstar + \sqrt{1-m^{2}}\bxi$, where $\bxi\sim\mathcal{N}(0,\idmat_{d})$ and $m = \sfrac{1}{d}\langle\wstar,\hat{\w}\rangle$ is the correlation with the target. Then:
\begin{align*}
    &\mathbb{P}\left[\exists\text{ consis. } \attack\in H_{\attnorm}(\varepsilon)\right] \\
    &\qquad=2\Phi\left(\varepsilon\sqrt{d}\frac{\sqrt{1-m^{2}} d^{\star}_{q^{\star}}(\bxi)}{\|\hat{\w}\|_{2}}\right)-1\,.
\end{align*}
Since $d^{\star}_{q^{\star}}(\bxi) = \Theta(d^{1/q^{\star}})$ as $d\to\infty$, for fixed $\varepsilon=\Theta(1)$ and $q>1$ the probability approaches one unless $m^{2}=1$ (perfect alignment). This highlights the vulnerability of high-dimensional predictors, though sparse predictors can exhibit different behavior since $d^{\star}_{q^{\star}}(\hat{\w}_{\perp})$ depends on the part of the predictor support that does not overlap with the target. We report how
the existence probability depends on the attack geometry $\attnorm$ and the dimension $d$ in~\Cref{fig:direct-space-prob} \textit{(Left, Center)}. \end{remark}
\begin{remark}[Algorithmic vs.\ statistical irreducibility]
\label{rmk:irreducibility}
The probability in~\Cref{rmk:Gaussian} is strictly positive whenever $m^{2}{<}1$, with two contributing sources. The \emph{algorithmic gap}, the suboptimality of a given estimator relative to the best achievable one, can be reduced by tuning the estimation of $\what$ to maximize $m$. The \emph{statistical gap} is the residual misalignment of the \emph{best possible} estimator at finite $\alpha$: for the continuous weight priors considered here, even the Bayes-optimal estimator achieves only $m^{2}<1$, with generalization error decaying as $\Theta(1/\alpha)$ but never vanishing~\citep{aubin_generalization_2020}. This contrasts with discrete or sparse priors, where a phase transition can yield exact recovery above a critical $\alpha$~\citep{barbier_optimal_2019}. Consistent attacks are thus an \emph{irreducible} statistical phenomenon under our assumptions.
\end{remark}

\begin{remark}\label{rmk:powerlaw}
Consider $\boldsymbol{\Sigma}$ with \emph{power law spectrum} with eigenvalues $\lambda_{i} \propto i^{-\beta}$ (for $\beta > 0$) normalized to $\sum_i \lambda_i {=}d$ and let $\{\mathbf{v}_i\}_{i=1}^d$ denote the corresponding orthonormal eigenbasis. Suppose $\wstar{=}\sqrt{d}\mathbf{v}_1$ and that the predictor has a fixed correlation $m$ with the target. As $\beta$ increases, the spectrum concentrates on $\lambda_1$ (which grows toward $d$), reducing the attack probability for all predictors (\Cref{fig:direct-space-prob}, \textit{Right}).
However, vulnerability also depends on where
$\hat{\w}_{\perp}$
distributes its mass in the eigenbasis, the two cases are:
\begin{itemize}
    \item \textbf{Low-rank (LR) alignment}: When $\hat{\w}_{\perp} \!\propto \!\sum_{i=2}^{k+1} \mathbf{v}_i$ concentrates on the top-$2{:}(k{+}1)$ eigenmodes and it benefits from eigenvalues $\lambda_2,\ldots,\lambda_{k+1}$ being large.
    \item \textbf{Tail alignment}: When $\hat{\w}_{\perp}\! \propto\! \sum_{i=k+2}^{d} \mathbf{v}_i$ concentrates on tail modes, it places norm on eigenmodes $\lambda_{k+2},\ldots,\lambda_d$.
\end{itemize}

\Cref{fig:direct-space-prob} (\textit{Right}) shows how the tail predictor is \emph{more vulnerable} than the LR predictor by a constant logit-level factor.
\end{remark}

\subsection{Robust Empirical Risk Minimization}
\label{sec:robust-training}
\revision{A natural question is whether robust training can effectively mitigate consistent attacks. Robust empirical risk minimization (ERM) learns robust classifier rules from a dataset $\mathcal{D} = \{(\x_\dataidx, y_\dataidx)\in\mathbb{R}^{d}\times\{-1,+1\}: \dataidx=1,\dots, n\}$.}
From
$\mathcal{D}$, the statistician estimates a classifier by optimizing the robust \textit{empirical} (regularized) risk, defined
\begin{equation}\label{eq:risk-definition}
    \mathcal{L} (\w) = \sum_{\dataidx = 1}^{\nsamples}
    \max_{
        \revision{\norm{\defend_\dataidx}_\defnorm \leq R}
    }
    \lossfun \qty(
        y_\dataidx \langle \w, \x_\dataidx + \defend_\dataidx \rangle
)
    + \lambda \norm{\w}_2^2 \,,
\end{equation}
where $\lossfun: \mathbb{R} \to \mathbb{R}_+$ is a non-increasing convex loss function
and $\lambda \geq 0$ is a regularization parameter, \revision{while $R\geq0$ is the training attack radius.}
\revision{The case $R=0$ corresponds to standard ERM, while $R>0$ corresponds to robust training.}
Given the dataset $\mathcal{D}$, we estimate the model's weights as
{\setlength{\belowdisplayskip}{2pt}\setlength{\abovedisplayskip}{2pt}
\begin{equation}\label{eq:risk-minimization}
    \hat{\w} \in \argmin_{\w \in \revision{\RR^d}} \mathcal{L} (\w) \,.
\end{equation}
}
\revision{While robust training has proven effective in practice, understanding its protection against consistent attacks still requires analysis.}
\subsection{High-Dimensional Asymptotic Analysis}
\label{sec:highd}
Motivated by~\Cref{rmk:Gaussian}, we now investigate the behavior of both standard and robustly trained predictors in the high-dimensional limit where consistent adversarial attacks proliferate. More concretely, we will derive sharp asymptotic results for the case where $\hat{\w}$ is a trained predictor under the Gaussian design assumption, and discuss the benefits of robust ERM in mitigating consistent adversarial attacks. We will work under the following assumptions.

\begin{assumption}[Data distribution]\label{asm:data-distr-well-spec}
We assume the covariates are isotropic Gaussian $\x \sim \mathcal{N}(\mathbf{0}, \sfrac{1}{d}\idmat_d)$ and that labels are generated from a ground-truth linear classifier $y\sim {\rm Rad}(f_{\w_{\star}}(\x))$ where $f_{\wstar}(\x) = \mathbb{P}(y=+1|\x)=\varphi(\langle \wstar,\x\rangle)$ with monotonic link function $\varphi$ and ground-truth weights $\wstar \in \mathbb{S}^{d-1}(\sqrt{d})$.
\end{assumption}

\begin{assumption}[Scaling of the Adversarial Strength]\label{asm:scaling-eps-well-spec}
For a given perturbation geometry $\attack\in B_{\attnorm}(\varepsilon)$ with $q>1$, we assume that $\varepsilon = O_d(d^{-\sfrac{1}{q^\star}})$ as $d\to\infty$, where $q^{\star}$ is the dual. We define the rescaled radius as
$\tilde{\varepsilon} = \varepsilon\, d^{\sfrac{1}{q^\star}}$.
\revision{Likewise, for the $\ell^s$ inner maximization in~\cref{eq:risk-definition}, we take $R=r\,d^{-1/s^\star}$, where the dimension-free training budget $r=O(1)$ is the parameter entering the asymptotic formulas.}
\end{assumption}

\begin{remark}
As briefly discussed in~\Cref{rmk:Gaussian},~\Cref{asm:scaling-eps-well-spec} provides the right scaling for non-trivial attacks in the high-dimensional limit considered in this work: a slower scaling would result in a perturbation strength which is too weak and any faster scaling would result in a perturbation that flips any label. The same scaling was considered in previous asymptotic analyses of robust training in \cite{taheri_asymptotic_2021,javanmard_precise_tradeoffs_2020,tanner2024high}.
\end{remark}

Robust adversarial training for well-specified linear classifiers on Gaussian covariates (\Cref{asm:data-distr-well-spec}) has been studied by \cite{vilucchio2024geometry} in the proportional high-dimensional asymptotics where $n$ diverges with $d$ at constant ratio $\alpha = \sfrac{n}{d}=\Theta(1)$.
They characterize the limiting distribution of the robust estimator $\what$ for general $\ell^s$-robust training through a nonlinear system of equations.

For clarity of exposition, we present our result about the limiting behavior of consistent metrics for $s=2$ and noiseless channel (for the generic form see~\Cref{sec:app:limiting-error-metrics}).

\begin{theorem}[Consistent metrics for well-specified model]\label{thm:metrics-well-spec}
\revision{Under~\Cref{asm:data-distr-well-spec,asm:scaling-eps-well-spec}, with a noiseless link,} the metrics defined in \cref{eq:def-misclassif-proper,eq:boundary-proper} with decision rule $\hat{y}(\x) = \sign(2 f_{\what} (\x) - 1)$, where $\what$ comes from \cref{eq:risk-minimization}, concentrate in the high-dimensional limit where $n,d\to \infty$ with $\sfrac{n}{d} = \Theta(1)$ to the following two dimensional integrals
{
\begin{align}
    \!\!\!\!\!\adverrproper \!&=\!\!\int\!\!\dd p(z_1,\! z_2)\,
    \mathbbm{1}\!\qty{\sign(z_1)z_2-\tilde{\varepsilon}\mathcal A<0} , \label{eq:well-specified-rob-proper} \\
    \!\!\!\!\!\bounderrproper \!&=\!\!\int\!\!\dd p(z_1,\! z_2)\,
    \mathbbm{1}\!\qty{0<\sign(z_1)z_2<\tilde{\varepsilon}\mathcal A} , \label{eq:well-specified-bnd-proper}
\end{align}
}
where $\mathcal{A}=\sqrt{2(\tau - m^2)} \sqrt[\attnorm^\star]{\Gamma\qty(\frac{\attnorm^\star + 1}{2}) / \sqrt{\pi}}$
and
$(z_1,z_2)\sim\mathcal{N}\qty(0, \qty(\begin{smallmatrix}
    1 & m \\
    m & \tau
\end{smallmatrix}))$ with
$\tau=\sfrac{1}{d}\|\what\|_2^2$ the predictor's $\ell_2$-norm
and $m=\sfrac{1}{d}\langle \wstar, \what\rangle$ its scalar product with the target.
\end{theorem}
The proof proceeds by explicitly solving the inner maximization and then showing concentration to the previous expression in the high-dimensional limit; see~\Cref{sec:app:limiting-error-metrics}.
The values of $\tau$ and $m$ can be obtained from the characterization of $\what$ in \cite{vilucchio2024geometry}.

\begin{remark}[Intuition via local fields]\label{rmk:intuition-well-spec}
\revision{\Cref{eq:well-specified-rob-proper,eq:well-specified-bnd-proper} admit a \emph{local fields} reading \citep{clarte_double_descent}: under the normalized design $\x\sim\mathcal N(0,\idmat_d/d)$, the Gaussian pair $(z_1,z_2)=(\langle \wstar,\x\rangle,\langle \what,\x\rangle)$ encodes the signed distances of $\x$ to the target and model boundaries. The attack subtracts $\tilde\varepsilon\mathcal A$ from the signed model margin $\sign(z_1)z_2$; hence the boundary error is precisely the probability that this initially positive margin lies between $0$ and $\tilde\varepsilon\mathcal A$. The attack surface $\mathcal{A}\propto\|\what_{\perp}\|_2$ only sees the predictor component orthogonal to $\wstar$ (\Cref{prop:exis}).}
\end{remark}

\begin{figure}
\centering
\hfill
    \includegraphics[width=0.45\linewidth]{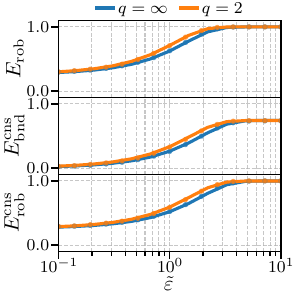}
    \hfill
    \includegraphics[width=0.45\linewidth]{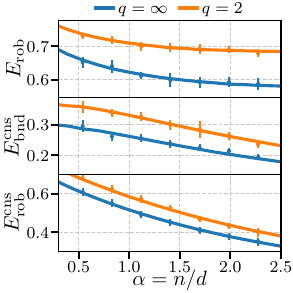}
\caption{
\textit{(Left)}
        Dependence of the metrics in \cref{eq:classical-robust-error,eq:def-misclassif-proper,eq:boundary-proper} for the well-specified model as a function of the attackers norm.
        Points are 10 different simulations for $d=500$, showing good agreement already at finite dimension.
        \textit{(Right)}
        Performance difference for optimally regularized non-robust training under attacks constrained with different norms.
        The points show 10 different simulations with $d = 500$ and $n$ scaled such that $n = \alpha d$.
    }
    \label{fig:direct-space}
\end{figure}

For completeness, we report the limiting value of $\advgenerr$ already derived in~\citep{tanner2024high} as being
\begin{equation*}{\begin{aligned}
    \advgenerr
    &=\int\dd p(z_1,z_2)\,
    \mathbbm{1}\!\qty{\sign(z_1)z_2-\tilde\varepsilon\mathcal B<0},\\[-2pt]
    \mathcal B
    &=\sqrt{2\tau}\sqrt[\attnorm^\star]{\frac{\Gamma\qty(\frac{\attnorm^\star+1}{2})}{\sqrt\pi}}\,.
    \end{aligned}}
\end{equation*}
\revision{The consistent versions of the errors (\cref{eq:well-specified-rob-proper,eq:well-specified-bnd-proper}) depend on the quantity $\sqrt{\tau - m^2}$} while
$\advgenerr$ depends on $\sqrt{\tau}$. Since $\sqrt{\tau - m^2} < \sqrt{\tau}$ (as $\tau \geq m^2 > 0$), this
explains why consistent attacks are less effective than inconsistent ones for the same attack strength $\varepsilon$, confirming our earlier result from~\Cref{cor:incon}
and as illustrated in~\Cref{fig:direct-space} \textit{(Left)}.
The former quantity is precisely the length of $P_{\perp}\what$ appearing in \Cref{prop:exis}.
\Cref{sec:app:additional-experiments} presents additional experiments comparing consistent and inconsistent boundary errors.

\Cref{fig:direct-space}~\textit{(Left)} shows the asymptotic dependence of the metrics in~\Cref{def:metrics} with the rescaled perturbation strength $\tilde{\varepsilon}$ in the high-dimensional limit. This provides a quantitative measure of how strong a consistent adversarial attack needs to be to flip a certain percentage of the classifier labels: for instance, to flip $50\%$ of the labels with an $\ell_2$-attack one needs a strength of $\tilde{\varepsilon}\approx 1$ ($\varepsilon \approx d^{-\sfrac{1}{2}}$).

Using the results from \cite{vilucchio2024geometry} to find the values of $\tau$ and $m$ we can use~\Cref{thm:metrics-well-spec} to analyze the consistent errors of the robustly trained $\what$ from~\cref{eq:risk-minimization} trained on $n$ samples in dimension $d$, where both $n$ and $d$ tend to infinity with fixed ratio $\alpha = \sfrac{n}{d}$.

\revision{\Cref{fig:direct-space} \textit{(Right)} shows the performance of robustly trained $\what$ as a function of $\alpha = \sfrac{n}{d}$, demonstrating a monotonic decrease of all metrics with the sample complexity $\alpha$ for two different attack geometries. While the errors $\advgenerr$ and $\adverrproper$ start from the same values, $\adverrproper$ decreases faster with $\alpha$, indicating that with more samples the model learns representations that are more effective against consistent adversarial attacks.}

We observe that $\ell^2$-attacks ($q\!=\!2$) yield
higher robust errors than $\ell^\infty$-attacks ($q\!=\!\infty$), demonstrating that attack geometry affects
vulnerability beyond the difference in perturbation budget available across norms: while~\Cref{asm:scaling-eps-well-spec} ensures attacks are non-trivial, the geometric properties of each norm determine how effectively that budget translates into successful adversarial perturbations.
Additional experiments are presented in~\Cref{sec:app:additional-experiments}.

\section{The Role of Overparameterization: Latent Variable Model}
\label{sec:latent-space-model}

Despite many empirical works on the subject, the interplay between adversarial attacks and overparameterization remains poorly understood, with contradictory evidence on the susceptibility of large neural networks to adversarial attacks \citep{chen2024over}.

In this section, we investigate this question on a popular mathematical testbed for studying the role of overparameterization, the \emph{latent variable model} \citep{hastie2022surprises}. In this model, the ground-truth classifier $f_{\wstar}(\z) = \varphi(\langle \wstar, \z \rangle)$ is defined in a latent space with latent covariates $\z \in \RR^d$ and weights $\wstar \in \mathbb{S}^{d-1}(\sqrt{d})$. Labels are generated according to the latent rule $y \sim {\rm Rad}(f_{\wstar}(\z))$ as in~\cref{eq:probit-def}. The statistician does not observe the latent covariates $\z\in\mathbb{R}^d$ directly but instead has access to a transformed representation $\x \in \mathbb{R}^p$ defined as $\x = \featmat \z + \uu$,
where $\featmat \in \mathbb{R}^{p \times d}$ is the \emph{feature matrix} and $\uu \sim \mathcal{N}(\boldsymbol{0}, \sfrac{1}{p}\idmat_p)$ is an independent
noise term.

While this model seems simplistic, recent Gaussian universality results have shown that in the proportional limit, ERM on this latent variable model is equivalent to ERM on a two-layer neural network with frozen first-layer weights (a.k.a. \emph{random features model}) \citep{mei_generalization_2022, Goldt_2021, Gerace_2021, universality_laws, montanari2022universality}.
This places the latent variable model as a convenient testbed to study the phenomenology associated to overparameterized networks --- such as benign interpolation --- in a mathematically tractable setting.
In this mapping, the level of overparameterization is given by the features dimension $p$. When $p > d$, we will say the model is \textit{overparameterized}, and when $p < d$, the model is \textit{underparameterized}.

\subsection{Geometry of Consistent Attacks on the Latent Space}
We now discuss the geometrical properties of consistent attacks in the latent variable model. Note that in this context an adversary could either attack the latent space ($\attack \in \RR^d$) or feature space ($\attack \in \RR^p$). Considering perturbations in feature space, \textit{i.e.} perturbations to $\x$, will result in a similar analysis as the one carried out for the model of~\Cref{sec:existence-proper-adversarial}.
\revision{Therefore, in the following we focus on the former, where the conditions in \Cref{def:consistent} translate to:}
\begin{itemize}
    \item \revision{\textbf{Target-score invariance:} $\attack\in H_{\attnorm}(\varepsilon) = \{\attack \in B_{\attnorm}(\varepsilon): \langle \wstar, \attack\rangle =0\}$.}
    \item \revision{\textbf{Model deception:} $|\langle\hat{\btheta},\featmat\attack\rangle|> |\langle \hat{\btheta}, \featmat\z + \uu\rangle|$ and $\sign(\langle\hat{\btheta},\featmat\attack\rangle)=-\sign(\langle \hat{\btheta}, \featmat\z + \uu\rangle)$.}
\end{itemize}
where the model weights are denoted by $\hat{\btheta}\in\mathbb{R}^{p}$ to avoid confusion. Adapting the argument in~\Cref{sec:geometry:wellsp} to this case is straightforward, leading to the following characterization of consistent latent space attacks.
\begin{proposition}[Existence of consistent latent space attacks]
\label{prop:exist:latent}
\revision{Consider the setting of binary classification in the latent space model: a linear classifier with weights $\wstar\in\mathbb{S}^{d-1}(\sqrt{d})$ assigns labels according to $y\sim {\rm Rad}(f_{\wstar}(\z))$, where $f_{\wstar}(\z) = \varphi_{\star}(\langle \wstar, \z\rangle)$, while the statistician observes only the pairs $(\x,y)\in\mathbb{R}^{p}\times \{-1,+1\}$ with $\x=\featmat\z+\uu\in\mathbb{R}^{p}$, fitting a linear classifier $f_{\thetahat}(\x)=\varphi(\langle \thetahat, \x\rangle)$ with weights $\thetahat\in\mathbb{R}^{p}$.} Then, a consistent attack $\attack\in B_{\attnorm}(\varepsilon)$ with respect to $\wstar\in\mathbb{S}^{d-1}(\sqrt{d})$, $\z\in\mathbb{R}^{d}$ and the decision function $\hat{y}(\x)=\sign(2f_{\thetahat}(\x)-1)$ exists if and only if:
{\setlength{\belowdisplayskip}{2pt}\setlength{\abovedisplayskip}{2pt}
\begin{align}
\label{eq:cond}
      \varepsilon d^{\star}_{q^{\star}}(P_{\perp}F^{\top}\hat{\btheta}) \geq |\langle \hat{\btheta},F\z+\uu\rangle |
\end{align}
}
where $P_{\perp}=\idmat_{d}-\sfrac{\wstar\wstar^{\top}}{d}$ is the projector in the space orthogonal target weights and $q^{\star}$ is the dual of $q$.
\end{proposition}
\begin{remark}
While in the well-specified case the vulnerable region is determined by the margin $\kappa_{\attnorm}(\x)$ in~\cref{eq:def:marginq}, in the latent model this is defined by the latent margin:
{\setlength{\belowdisplayskip}{2pt}\setlength{\abovedisplayskip}{2pt}
\begin{align}
    \eta_{\attnorm}(\z) \coloneqq \frac{|\langle\hat{\btheta},\featmat \z+\uu\rangle|}{d^{\star}_{q^{\star}}(P_{\perp}F^{\top}\hat{\btheta})} \,.
\end{align}
}
Note that this can be larger or smaller than $\kappa_{\attnorm}(\x)$, depending on the details of the problem.
\end{remark}
\begin{corollary}[Existence for Gaussian latent variables]\label{corr:prob-latent}
In the case of i.i.d.~Gaussian latent variables $\z\!\sim\!\mathcal{N}(0,\sfrac{1}{d}\idmat_{d})$ and $\uu\sim\mathcal{N}(0,\sfrac{1}{p}\idmat_{p})$, the probability a consistent attack exists is given by:
{\setlength{\belowdisplayskip}{2pt}\setlength{\abovedisplayskip}{2pt}
\scriptsize
\begin{align*}
\mathbb{P}\left[\exists\text{ consistent } \attack\in \!H_{\attnorm}(\varepsilon)\right] \!=\! 2\Phi\!\left(\frac{\sqrt{p}\varepsilon d^{\star}_{q^{\star}}(P_{\perp}F^{\top}\hat{\btheta})}{\sqrt{||\hat{\btheta}||^{2}_{2}+\sfrac{p}{d}||\featmat^{\top}\hat{\btheta}||^{2}_{2}}}\right)\!-\!1,
\end{align*}
}
where $\Phi$ is the standard normal c.d.f.
\end{corollary}
\begin{remark}
    \revision{In the latent variable model, the probability of existence depends on the predictor's latent-space projection $\featmat^{\top}\hat{\btheta}$ rather than on the predictor alone. In particular, components of $\hat{\btheta}\in\mathbb{R}^{p}$ lying in ${\rm Ker}(\featmat^{\top})$ contribute only to the $\ell^{2}$-norm in the denominator. Thus, in the overparameterized setting $p>d$, one can reduce the probability of a consistent attack either by increasing alignment with the target $\langle \wstar, \featmat^{\top}\hat{\btheta}\rangle$ or by concentrating most of the norm in the $p-d$ directions in ${\rm Ker}(\featmat^{\top})$.} This is closely related to the conditions for benign overfitting in \cite{bartlett2020benign}.
\end{remark}

The consistent and inconsistent adversarial errors associated to a predictor $\hat{y}$ in the latent space model are
{\small\begin{align}
    \!\!\!\!\!\adverrproper \!&=\!
    \mathbb{E}
\qty[
        \!\max_{
            \substack{
                \attack\in B_{\attnorm}(\varepsilon): \\
                \revision{\langle\wstar,\z\rangle=\langle\wstar,\z+\attack\rangle}
            }
        } \!\!\!\!\!\!\!
        \mathbbm{1}\qty{
            y
            \neq
            \hat{y}(\featmat(\z + \attack) + \uu)
        }
    ] , \label{eq:def-misclass-proper-latent-space} \\
    \!\!\!\!\!\bounderrproper \!&=\! \mathbb{E}
\!\qty[
        \max_{
            \substack{
                \attack\in B_{\attnorm}(\varepsilon): \\
                \revision{\langle\wstar,\z\rangle=\langle\wstar,\z+\attack\rangle}
            }
        } \!\!\!\!\!\!\!
        \mathbbm{1}\qty{
        \genfrac{}{}{0pt}{}{
            y
            \neq
            \hat{y}(\featmat(\z + \attack) + \uu)
        }{
            y = \hat{y}(\featmat \z + \uu)
        }
        }
] , \label{eq:def-boundary-proper-latent-space} \\
    \!\!\!\!\!\advgenerr \!&=\! \mathbb{E}
\qty[
        \max_{\attack\in B_{\attnorm}(\varepsilon)} \!\!
        \mathbbm{1}\qty{
            y
            \neq
            \hat{y}(\featmat(\z + \attack) + \uu)
        }
    ] . \label{eq:def-adverr-improper-latent-space}
\end{align}}

\subsection{High-Dimensional Asymptotics}
We now move to the analysis of trained predictors in the context of the latent variable model.
Consider training data $\mathcal{D}=\{(\x_\dataidx, y_\dataidx)\in\mathbb{R}^{p}\times \{-1,+1\}: \dataidx=1,\dots n\}$ independently drawn from the latent variable model. Our goal in this section is to characterize the asymptotic behavior of the estimated binary linear classifier defined by the vector $\thetahat$ estimated from $\mathcal{D}$ using~\cref{eq:risk-definition,eq:risk-minimization}.\footnote{\revision{Here the observed dimension is $p$, so~\cref{eq:risk-definition,eq:risk-minimization} are read with $\w\in\mathbb{R}^{p}$ and $R=r\,p^{-1/s^\star}$.}} Our results will hold under the following assumptions.
\begin{assumption}[High-dimensional limit]\label{asm:high-dimensional-limit}
We consider the proportional high-dimensional limit where $n,p,d\to\infty$ at fixed ratios $\alpha := \sfrac{n}{d}$ and $\psi := \sfrac{p}{n}$. For convenience, we also define $\gamma := (\alpha\psi)^{-1} = \sfrac{d}{p}$.
\end{assumption}

{
\begin{assumption}[Data Distribution Latent Space Model]\label{asm:data-distr-latent-space}
We assume that data $(\x,y)\in\mathbb{R}^{p}\times \{-1,+1\}$ is drawn from a latent variable model with $\z \sim \mathcal{N}(\mathbf{0}, \idmat_d/d)$ and ground-truth linear classifier $f_{\wstar}(\z) = \varphi(\langle \wstar,\z\rangle)$ with $\wstar \in\mathbb{S}^{d-1}(\sqrt{d})$. The observed features $\x\in\mathbb{R}^{p}$ are generated as $\x = \featmat \z + \uu$ with independent noise $\uu \sim \mathcal{N}(\mathbf{0}, \idmat_p/p)$ and
{\small
\begin{equation}\label{eq:feature-mat-definition}
    \featmat = \begin{cases}{
        \Bigg[\begin{array}{c}
        \idmat_d \\
        \mathbf{0}_{(p-d) \times d}
        \end{array}\Bigg]} & \text { if } p \geq d \\
        {\sqrt{\frac{d}{p}}\big[\begin{array}{ll}
            \idmat_p & \mathbf{0}_{p \times(d-p)}
        \end{array}\big]} & \text { if } p<d
    \end{cases}\,.
\end{equation}
}
\end{assumption}
This normalization keeps both fields dimension-free: $\operatorname{Var}(\langle\wstar,\z\rangle)=1$ and, conditionally on $\thetahat$, $\operatorname{Var}(\langle\thetahat,\x\rangle)=\|\featmat^\top\thetahat\|_2^2/d+\|\thetahat\|_2^2/p$.
}

\begin{remark}
All the phenomenology that follows also holds for a random Gaussian feature matrix $F\in\mathbb{R}^{p\times d}$.
The choice of feature matrix in~\cref{eq:feature-mat-definition} was previously considered in \cite{hastie2022surprises} in the context of ridge regression.
\Cref{thm:self-consistent-equations-latent-space} extends the discussion to binary classification.
\end{remark}

\revision{The main technical result for this part consists in characterizing the high-dimensional behavior of the robust empirical risk minimizer $\thetahat$ as the solution of a system of self-consistent equations.}
For clarity of exposition we state here the result for $s = 2$ in~\cref{eq:risk-minimization,eq:risk-definition} and leave the generic statement to~\Cref{sec:app:proof-high-d}.

\begin{theorem}[\revision{Self-Consistent Equations for the Latent Space Model}]\label{thm:self-consistent-equations-latent-space}
Under~\Cref{asm:data-distr-latent-space,asm:high-dimensional-limit,asm:scaling-eps-well-spec} the values of
$m \!=\! \frac{\wstar^\top \featmat^\top \thetahat}{d} $, $\tau \!=\! \frac{\|\featmat^\top \thetahat\|_2^2}{d} \!+\! \frac{\|\thetahat\|_2^2}{p} $ and $P \!=\! \frac{\| \thetahat \|_2^2}{p}$
concentrate in high dimension to the solution of the following system of self-consistent equations. The self-consistent equations are made of a first set
\begin{equation}\label{eq:latent-space-eq-channel}
    \begin{cases}
        \bar{m} = \alpha \sqrt{\gamma} \mathbb{E}_{\xi\sim \mathcal{N}(0,1)}\left[
            \int_{\RR} \dd{y}
            \partial_\omega \mathcal{Z}_0
            f_\lossfun
        \right] \\
        \bar{\tau} = \alpha \gamma \mathbb{E}_{\xi\sim \mathcal{N}(0,1)}\left[
            \int_{\RR} \dd{y}
            \revision{\mathcal{Z}_0
            f_\lossfun^2}
        \right] \\
        \bar{V} = -\alpha \gamma \mathbb{E}_{\xi\sim \mathcal{N}(0,1)}\left[
            \int_{\RR} \dd{y}
            \revision{\mathcal{Z}_0
            \partial_\omega f_\lossfun}
        \right] \\
        \bar{P} = 2 r P^{\sfrac{1}{\defnorm}} \mathbb{E}_{\xi\sim \mathcal{N}(0,1)}\left[
            \int_{\RR} \dd{y}
            \revision{y\,\mathcal{Z}_0
            f_\lossfun}
        \right]
    \end{cases}\,,
\end{equation}
that depend on the loss function $\ell$
through $\mathcal{Z}_0 \equiv \mathcal{Z}_0(y, \sfrac{m}{\sqrt{\tau}} \xi, 1 - \sfrac{m^2}{\tau})$ and $f_\lossfun \equiv f_\lossfun(y, \sqrt{\tau} \xi, V, P)$, where
$\mathcal{Z}_{0}(y, \omega, V) = \mathbb{E}_{z \sim \mathcal{N}(\omega,V)}\qty[ \mathbb{P}(y \mid z) ]$ and
\revision{$f_{\lossfun}(y, \omega, V, P) = (\mathcal{P}_{V \lossfun(y, \cdot - y r \sqrt[\defnorm^\star]{P}) } \qty(\omega) - \omega) / V$.}

The second set of equations for $\gamma \leq 1$ is
\begin{equation}\label{eq:latent-space-eq-prior-gamma-small}
    \!\!\!\!\!\!\!\begin{cases}
        m \!=\! \frac{\bar{m}}{2\lambda + \bar{P} + \bar{V}\qty(1 + \frac{1}{\gamma})}  \\
        \tau \!=\! \frac{1}{(2\lambda + \bar{P} + \bar{V}\qty(1 + \frac{1}{\gamma}))^2} \! \qty[
            \bar{m}^2 \frac{1-\gamma}{\gamma} \!+\! \bar{\tau}\frac{(1+\gamma)^2 + \gamma - \gamma^2}{\gamma}
        ] \\
        V \!=\! \frac{1}{2\lambda + \bar{P} + \bar{V}\qty(1 + \frac{1}{\gamma})}
        \\
        P \!=\! \frac{1}{(2\lambda + \bar{P} + \bar{V}\qty(1 + \frac{1}{\gamma}))^2} \qty[\bar{m}^2 + 2\bar{\tau}]
    \end{cases}
\end{equation}
and for $\gamma > 1$ is
\begin{equation}\label{eq:latent-space-eq-prior-gamma-big}
    \begin{cases}
        m \!=\! \frac{\bar{m}}{2\lambda + \bar{P} + 2\bar{V}\vphantom{X^X}} \\
        \tau \!=\! \frac{\bar{m}^2 + 2\bar{\tau}}{(2\lambda + \bar{P} + 2\bar{V})^2 \vphantom{X^X}}
\end{cases}
    \begin{cases}
V \!=\! \frac{1}{2\lambda + \bar{P} + 2\bar{V}\vphantom{X^X}} \\
        P \!=\! \frac{\bar{m}^2 + 2\bar{\tau}}{(2\lambda + \bar{P} + 2\bar{V})^2\vphantom{X^X}}
    \end{cases}
\end{equation}

\end{theorem}

The proof (see~\Cref{sec:app:proof-high-d}) uses the Gordon Min-Max theorem to characterize the minimizer of the robust risk in~\cref{eq:risk-definition} via a low-dimensional set of self-consistent equations.
Such asymptotic characterizations via implicit equation systems are standard in the analysis of high-dimensional systems \citep{goldt_gaussian_2021, gerace2023gaussian, cui_error_2022, vilucchio2025asymptotics}.

With the previous theorem, we can characterize the
behavior of the proper adversarial errors in this data model.

\begin{theorem}[Proper Metrics for Latent Space Model]\label{thm:metrics-latent-space}
\revision{Under the same setting as~\Cref{thm:self-consistent-equations-latent-space}, with a noiseless link,} the metrics defined in~\cref{eq:def-misclass-proper-latent-space,eq:def-boundary-proper-latent-space} evaluated for $\thetahat$ from~\cref{eq:risk-minimization} and decision rule $\hat{y} = \sign(2 f_{\thetahat}(\x) - 1) $ concentrate to
{{
\begin{align}
    \!\!\!\!\!\adverrproper \!&=\!\!\int\!\!\dd p(z_1,z_2)\,
    \mathbbm{1}\!\qty{\sign(z_1)z_2-\tilde{\varepsilon}\mathcal C<0} \,, \label{eq:latent-space-rob-proper} \\
    \!\!\!\!\!\bounderrproper \!&=\!\!\int\!\!\dd p(z_1,z_2)\,
    \mathbbm{1}\!\qty{0<\sign(z_1)z_2<\tilde{\varepsilon}\mathcal C} \,, \label{eq:latent-space-bnd-proper}
\end{align}
}
}
where $\mathcal{C}$ is a complicated expression presented in~\Cref{sec:app:limiting-error-metrics}
and $(z_1,z_2) \sim \mathcal{N}\qty(0, \qty(\begin{smallmatrix}
    1 & m \\
    m & \tau
\end{smallmatrix}))$ with $\tau,m$ the solutions of the system of equations in~\Cref{thm:self-consistent-equations-latent-space}.
\end{theorem}
\begin{remark}
The difference between the two settings appears in the constant $\mathcal{C}$, which depends on the distribution of $\thetahat$ through the parameters found by the equations in~\Cref{thm:self-consistent-equations-latent-space}. Consequently, the local-fields interpretation of~\Cref{rmk:intuition-well-spec} carries over, with $\mathcal{C}$ now playing the role of the available attack surface $\mathcal{A}$.
\end{remark}
\begin{figure}
    \centering
    \includegraphics[width=0.49\linewidth]{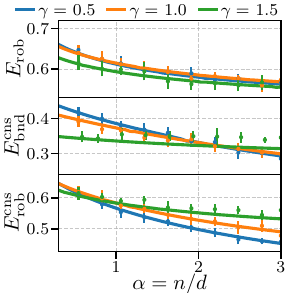}
\includegraphics[width=0.49\linewidth]{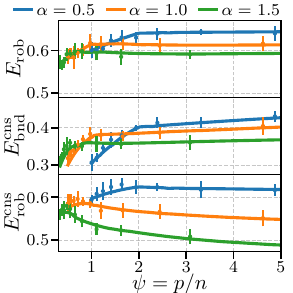}
\caption{
Error as a function of $\alpha$ and $\psi$ for the latent space model defined in~\Cref{sec:latent-space-model}.
        For both panels the lines are the exact asymptotic solution of~\cref{eq:latent-space-eq-prior-gamma-small,eq:latent-space-eq-prior-gamma-big,eq:latent-space-eq-channel} and the error bars are average and std over 10 realizations with $d=500$ and $p,n$ scaled accordingly.
        \textit{(Left)}
        \revision{Robust error as a function of the number of training samples. All metrics decrease as the amount of training data increases.}
        \textit{(Right)}
        \revision{Robust errors as a function of the number of latent-space parameters. We see that $\advgenerr$ and $\bounderrproper$ increase with the number of features, while $\adverrproper$ decreases.}
    }
    \label{fig:latent-space-sweeps}
\vspace{-1.5em}
\end{figure}

\subsection{The Interplay Between Overparameterization and Consistent Attacks}\label{sec:latent-space-overparam}
\revision{\Cref{thm:self-consistent-equations-latent-space,thm:metrics-latent-space} allow us to investigate the efficacy of consistent attacks on overparameterized models.}
We start by considering robust training with optimally tuned $\lambda$ and $r$. \revision{All three metrics considered in this work decrease as a function of the amount of training data $\alpha$; see~\Cref{fig:latent-space-sweeps} \textit{(Left)}. Thus, more data are always beneficial.} \revision{Interestingly, the curves cross for different overparameterization levels $\gamma$:}
\revision{no single level of overparameterization is optimal for every amount of available data.}

In~\Cref{fig:latent-space-sweeps} \textit{(Right)} we consider the role of optimal robust training as a function of different overparameterization levels $\psi = \sfrac{p}{n}$.
The three metrics
present a different behavior for large $\psi$.
The first two stay
approximately constant, with at most a mild increase,
\revision{while the third decreases with overparameterization, and this decrease becomes faster as more data are given to the model (larger $\alpha$).}

\revision{We study the behavior of the novel metrics $\bounderrproper$ and $\adverrproper$ in the regime of extreme overparameterization.}
{
\begin{proposition}[Extreme overparameterization]\label{prop:extreme-overparam}
Consider the setting of \Cref{thm:self-consistent-equations-latent-space,thm:metrics-latent-space} with logistic loss, a noiseless link, and $s{=}q{=}2$. 
As $\psi=p/n\to\infty$ (equivalently $\gamma=d/p\to0^+$ at fixed $\alpha$),
$m=m_0\sqrt\gamma+O(\gamma)$, 
$\tau=\tau_0+O(\sqrt\gamma)$, $P=O(\gamma)$,
where $m_0$ and $\tau_0>0$ are determined by the limiting saddle-point system. Then
{\small
\begin{align}
\partial_\psi\adverrproper
&=\frac{\alpha m_0}{2\pi\sqrt{\tau_0}}
e^{-\tilde\varepsilon^2/2}\gamma^{3/2}+O(\gamma^2)>0,\label{eq:extreme-rob-derivative}\\
\partial_\psi\bounderrproper
&=-\frac{\alpha m_0}{2\pi\sqrt{\tau_0}}
\left(1-e^{-\tilde\varepsilon^2/2}\right)\gamma^{3/2}+O(\gamma^2),\label{eq:extreme-bnd-derivative}
\end{align}}
so the leading coefficient in~\cref{eq:extreme-rob-derivative} (resp. \cref{eq:extreme-bnd-derivative}) is non-negative (resp. non-positive) for $\tilde\varepsilon, m_0\geq 0$.
\end{proposition}
}

\revision{Thus the finite-range trends in~\Cref{fig:latent-space-sweeps} (\textit{Right}) eventually turn over: in the extreme regime $\adverrproper$ increases, whereas $\bounderrproper$ decreases with the explicit coefficient in~\cref{eq:extreme-bnd-derivative}. This regime dependence may help explain apparently contradictory observations~\citep{chen2024over}; see~\Cref{sec:app:extreme-overparam} for the proof.}
We further verify that this dual effect is not an artifact of the linear estimator: replacing it with a trained two-layer ReLU network on the same data-generating process reproduces the same qualitative trends (\Cref{sec:app:nn-experiment}).
Additional experiments in~\Cref{sec:app:additional-experiments} include ablations of other latent space models \citep{Goldt_2021} and results on MNIST/FashionMNIST \citep{deng2012mnist,xiao2017fashion}.

\section{Conclusions}
\label{sec:conclusions}
We investigated the fundamental distinction between consistent and inconsistent adversarial attacks in high-dimensional binary classification, introducing the consistent robust error $\adverrproper$ and the consistent boundary error $\bounderrproper$ as refined metrics, and deriving their exact asymptotics for robust empirical risk minimization in both the well-specified and the latent variable settings.
\revision{Curiously, overparameterization has a regime-dependent effect on consistent robustness. Over the finite range in~\Cref{fig:latent-space-sweeps}, the boundary and overall errors follow opposite trends, while the extreme-limit expansion predicts an eventual turnover of both. This non-monotonicity may help explain contradictory empirical observations~\citep{chen2024over}.}

We hope these findings, though derived from exactly solvable models, prove valuable for the broader robust machine learning community. Rather than viewing overparameterization as inherently detrimental to adversarial robustness,
one should account for the specific robustness objective at hand, as overparameterization can improve overall performance. Moreover, our characterizations provide theoretical guidance for selecting regularization parameters and attack
budgets during robust training.
\paragraph{Limitations and future work.}
Our analysis is set in the proportional high-dimensional regime with Gaussian covariates; while this regime belongs to a broad universality class, relaxing it is an interesting direction. The latent variable model captures the random-features/NTK regime but not feature learning; extending the analysis beyond it is an important next step.
Finally, target-score invariance ($g_{\star}(\x+\attack)=g_{\star}(\x)$) is stronger than label preservation for non-injective links, so our metrics provide conservative lower bounds on label-preserving vulnerability; extending to approximate score invariance ($|g_{\star}(\x+\attack)-g_{\star}(\x)|<\eta$) is a natural generalization.

\vspace{-0.25em}
\section*{Acknowledgements}
We thank the anonymous reviewers for their constructive feedback, and Fanny Yang and Antonio Ribeiro for inspiring discussions. This work was supported by the Swiss National Science Foundation (SMArtNet 212049; OperaGOST 200390) and the French ANR France 2030 program (ANR-23-IACL-0008), under the Choose France--CNRS AI Rising Talents program.

\paragraph{Impact Statement.}
This paper presents work whose goal is to advance the field of Machine Learning. There are many potential societal consequences of our work, none of which we feel must be specifically highlighted here.

\bibliography{refs}
\bibliographystyle{icml2026}

\newpage
\appendix
\onecolumn

\section{Further Related Works}\label{sec:app:further-rel-works}

\paragraph{Exact Asymptotics:}
Our analysis builds upon the previous literature characterizing the properties of predictors in the high-dimensional proportional regime. This approach spans multiple theoretical frameworks: high-dimensional probability theory \citep{thrampoulidis2015gaussian,pmlr-v40-Thrampoulidis15,vilucchio2025asymptotics}, statistical physics approaches
\citep{mignacco2020role,Gerace_2021,bordelon20a,loureiro2021learning,pmlr-v206-okajima23a,adomaityte_classification_2023,adomaityte2023highdimensional}, and random matrix theory \citep{doi:10.1073/pnas.1307845110, 8683376, NEURIPS2020_a03fa308, mei_generalization_2022, xiao2022precise, pmlr-v202-schroder23a, schroder24a, Defilippis2024, tabanelli2025computational}.
Our work is particularly motivated by recent advances in Gaussian universality \citep{Goldt_2021,montanari2022universality,dandi2023universality}, which demonstrate that simple Gaussian models often provide surprisingly accurate predictions for more complex data distributions in high-dimensions. This phenomenon emerges from concentration properties in high-dimensional spaces, leading to universality in generalization behavior across different covariate distributions \citep{tao2010random, donoho2009observed, pmlr-v162-wei22a, dudeja2023universality}.
Our technical contribution extends these frameworks to characterize consistent adversarial attacks: while existing asymptotic analyses focus on standard generalization or robust error, we derive the first closed-form high-dimensional limits for metrics that isolate label-preserving perturbations, revealing geometric dependencies on $\sqrt{\tau - m^2}$ rather than $\sqrt{\tau}$ alone.

\paragraph{Adversarial Robustness:}
Robust empirical risk minimization, commonly known as adversarial training, was pioneered in computer vision \citep{GSS15,Mad+18} and has since evolved into a primary defense against adversarial attacks. Researchers have developed numerous approaches to improve its computational efficiency \citep{Sha+19b,RWK20} and statistical properties \citep{Zha+19,Che+21,Wan+23}. On the theoretical front, several works have investigated the properties of robust empirical risk minimization for linear models \citep{raghunathan2020understanding, dan2020sharp, NEURIPS2023_4aa13186, pmlr-v258-ribeiro25a}, including sharp proportional asymptotics under different data designs \citep{javanmard_precise_tradeoffs_2020, HaJa22, Donhauser2021, dohmatobprecise, ribeiro2023overparameterized, tanner2024high,peng2026damage}.
A common thread in most of the previous analyses is the characterization of the standard robust error $\advgenerr$, treating all perturbations in $B_q(\varepsilon)$ uniformly without distinguishing whether they preserve or alter the ground-truth label. For instance, \citet{taheri_asymptotic_2021} and \citet{javanmard_precise_statistical_2022} derive sharp asymptotics for $\advgenerr$ under adversarial training in well-specified linear models, while \citet{clarysse2022adversarial} study adversarial perturbations directed along the signal in a noiseless setting, and \citet{tanner2024high} extend the analysis to Mahalanobis-norm attacks.
\revision{In high-dimensional linear regression with missing data, \citet{xing2025missing} analyze ridge and $\ell_2$ adversarial training and develop a fast leave-one-out cross-validation approximation.}
\revision{Relatedly, \citet{Tsi+24} study how the implicit bias induced by optimization algorithms and architectures in robust ERM affects standard robust generalization. Their focus is on matching implicit regularization to the threat model, whereas we isolate label-preserving attacks and characterize their high-dimensional behavior.}
Our work departs from this line by considering the consistency constraint $\langle \wstar, \attack \rangle = 0$, which requires a fundamentally different geometric analysis: the existence and strength of consistent attacks depend on the predictor component orthogonal to the target (\Cref{prop:exis} and \Cref{lemm:adv-decreasing-fun}), a quantity that plays no role in unconstrained robust error.
Of particular relevance is \citet{vilucchio2024geometry}, which derives high-dimensional asymptotics for binary classification in the well-specified model and which we build upon in \Cref{sec:wellspe}. Our contributions beyond their work are the geometric existence conditions (\Cref{prop:exis}), the entire latent variable model analysis (\Cref{sec:latent-space-model}), generalizing previous results \citep{mei_generalization_2022,Goldt_2021}, and a characterization of how feature dimension $p$ affects consistent robustness differently than standard robust error (\Cref{sec:latent-space-overparam}).
Similarly, \citet{Donhauser2021} study the robust error of non-robustly trained $\ell^2$-regularized estimators in well-specified models, but do not consider adversarial training or the latent variable model.
\revision{Beyond the proportional regime, \citet{lee2026toward} draw a related distinction in a feature-learning analysis of adversarial distillation, separating failures on intrinsically hard-to-learn samples from robustness loss on otherwise correctly classified points, mirroring our distinction between failures due to label-preserving perturbations and those due to limited expressivity.}

\paragraph{Consistent Attacks:}
The idea that adversarial attacks should be imperceptible to some metric of interest (e.g.\ the human eyes in vision) underlies most of the empirical literature \citep{szegedy2013intriguing, GSS15, fort2024ensemble}. The notion of a consistent attack in the theoretical literature was explored in \citet{raghunathan2020understanding, Donhauser2021, NEURIPS2023_96189e90}.
These works establish that proper adversarial examples---perturbations preserving the Bayes-optimal label---can hurt robust generalization and that interpolating models may be vulnerable even without label noise. However, they do not provide: (i)~geometric characterizations of when consistent attacks \emph{exist} in high dimensions as a function of model-target alignment and attack geometry (our~\Cref{prop:exis,prop:exist:latent,corr:prob-latent,cor:incon}); (ii)~separate quantification of vulnerability on correctly vs.\ incorrectly classified points (our $E^\text{cns}_\text{bnd}$ metric, \cref{eq:boundary-proper}); or (iii)~exact asymptotic formulas under Gaussian designs revealing how overparameterization affects consistent robustness (our~\Cref{thm:metrics-latent-space,thm:metrics-well-spec,thm:self-consistent-equations-latent-space}). Our metrics thus enable fine-grained analysis distinguishing model failures due to label-preserving perturbations from those due to limited expressivity or sample size, providing precise predictions for the vulnerability landscape in high-dimensional linear classification.

\section{Additional Experimental Details}\label{sec:app:additional-experiments-setting}

\subsection{Setting of the Figures in the Main Text}

\revision{We jointly optimize the dimension-free training budget $r\geq0$ and ridge parameter $\lambda\geq0$ using the asymptotic theory, which gives a deterministic objective for each pair $(r,\lambda)$. We use the gradient-free Nelder--Mead algorithm implemented in \texttt{scipy.optimize.minimize}.}

\begin{description}
\item[\Cref{fig:direct-space-prob} \textit{(Left)}]
\revision{For each value of $q$, the empirical points are obtained by generating $n=1000$ samples in $d=10$ dimensions, with the model parameter set to $m=0.5$. The probability is the fraction of samples for which an adversarial perturbation of norm $\varepsilon$ exists, while the solid theory lines are computed using~\cref{eq:probGauss-well-spec-general}.}

\item[\Cref{fig:direct-space-prob} \textit{(Center)}]
Now we fix $q=2$ and vary the data dimension $d$.
The empirical points are again obtained by generating $n=1000$ samples for each $d$, with $m=0.5$; the probabilities and theoretical lines are computed as before.

\item[\Cref{fig:direct-space-prob} \textit{(Right)}]
We consider Gaussian covariates with diagonal covariance $\boldsymbol{\Sigma}=\mathrm{diag}(\lambda_1,\ldots,\lambda_d)$ and power-law spectrum $\lambda_i \propto i^{-\beta}$, normalized so that $\sum_{i=1}^d \lambda_i=d$, and sweep $\beta$.
We fix the target $\wstar=\sqrt{d}\,\mathbf{e}_1$ (top eigenvector) and correlation $m=0.5$.
For each $\beta$, we construct two predictors: low-rank alignment with $\hat{\w}_{\perp}$ supported on the modes from $2$ to $k+1$ and tail alignment with support on modes $k{+}2{:}d$.
We set $\varepsilon=1$, $\attnorm = 2$, $k=10$, and evaluate $d\in\{16,128,1024\}$.
Solid curves plot~\cref{eq:probGauss-well-spec-general}; empirical points report the fraction of $n=10^{4}$ samples from $\mathcal{N}(\boldsymbol{0}, \boldsymbol{\Sigma})$ satisfying~\cref{eq:cond-well-spec}.

\item[\Cref{fig:direct-space} \textit{(Left)}]
\revision{The curves use $\w$ obtained from standard training, i.e. minimization of~\cref{eq:risk-minimization} with $r=0$, $\lambda=10^{-3}$, and training norm $s=2$. The sample complexity is fixed at $\alpha=1$.}
The points are produced as 10 different realizations with $d=500$ fixed.

\item[\Cref{fig:direct-space} \textit{(Right)}]
\revision{Here we show performance under evaluation attacks constrained in either $\ell^\infty$ or $\ell^2$. In both cases, $r$ and $\lambda$ are jointly tuned to minimize the corresponding error; the regularizer is ridge and the adversarial-training norm is $s=2$.}
The points are produced as 10 different realizations with $d=500$ fixed.

\item[\Cref{fig:latent-space-sweeps} \textit{(Left,Right)} ]
\revision{We consider robust ERM with training norm $s=\infty$, jointly tuning $r$ and $\lambda$ for each setting. The points are produced from 10 realizations with $d=500$ and $n,p$ scaled accordingly.}

\end{description}

\subsection{Additional Experiments}\label{sec:app:additional-experiments}

\paragraph{Alternative latent-space model}
To test whether our findings depend on the feature map in~\Cref{asm:data-distr-latent-space}, which generates the observed input $\x$ from the latent variable $\z$, we consider an alternative latent-space model.
\revision{Another model used to characterize overparameterization is the hidden manifold model \cite{goldt_gaussian_2021,Gerace_2021}, where the latent-space covariates are Gaussian, $\z \sim\mathcal{N}(\boldsymbol{0}, \idmat_d)$, and the feature-space covariates are a linear transformation $\x = \featmat \z$ with independent entries $\featmat_{ij} \sim \mathcal{N}(0,1)$.}

\begin{figure}[t]
    \centering
    \includegraphics[width=0.8\textwidth]{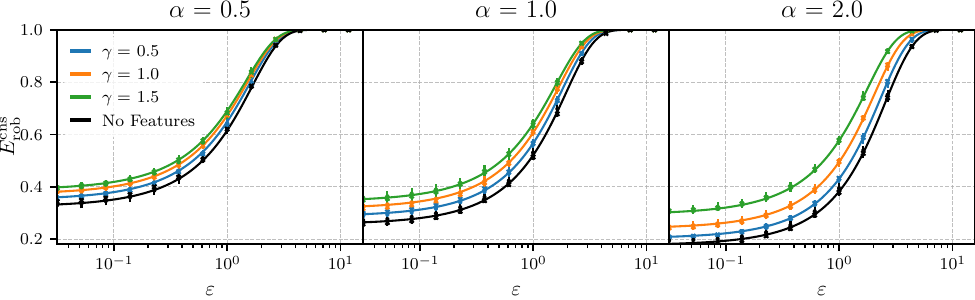}
    \caption{
        \revision{Behavior of $\adverrproper$ for Gaussian features as a function of the attack strength. The results use a model trained as in~\cref{eq:risk-definition,eq:risk-minimization}, with $\alpha$ and $\gamma$ specified in the figure.}
        The error bars refer to 10 repetitions of the experiments for $d=1024$.
        The metrics consider the $\what$ trained with $\lambda = 10^{-3}$, $r = 0.0$ and $\defnorm = \infty$.
    }
    \label{fig:app:dependance-overparam-robust-proper}
\end{figure}

\begin{figure}
    \centering
    \includegraphics[width=0.8\textwidth]{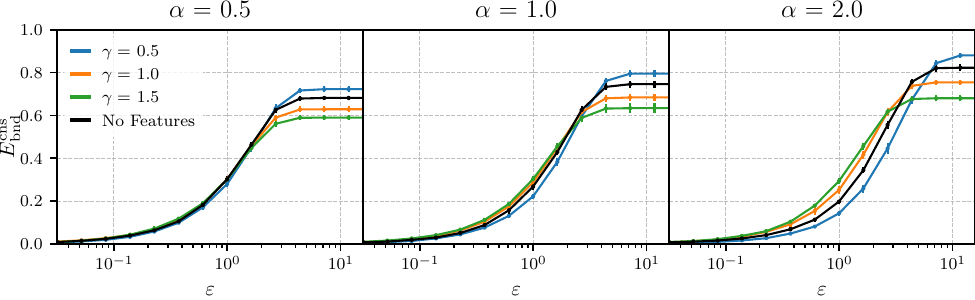}
    \caption{
        \revision{Behavior of $\bounderrproper$ for Gaussian features as a function of the attack strength. The results use a model trained as in~\cref{eq:risk-definition,eq:risk-minimization}, with $\alpha$ and $\gamma$ specified in the figure. The error bars refer to 10 repetitions with $d=1024$.}
        The metrics consider the $\what$ trained with $\lambda = 10^{-3}$, $r = 0.0$ and $\defnorm = \infty$.
    }
    \label{fig:app:dependance-overparam-boundary-proper}
\end{figure}

\begin{figure}[t]
    \centering
    \hfill
    \includegraphics[width=0.28\textwidth]{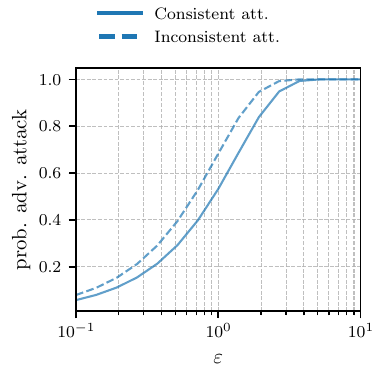}
    \hfill
    \includegraphics[width=0.28\textwidth]{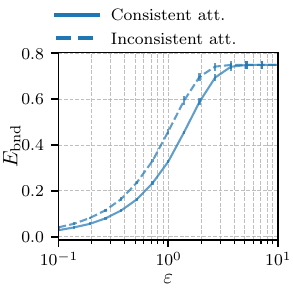}
    \hfill
    \includegraphics[width=0.28\textwidth]{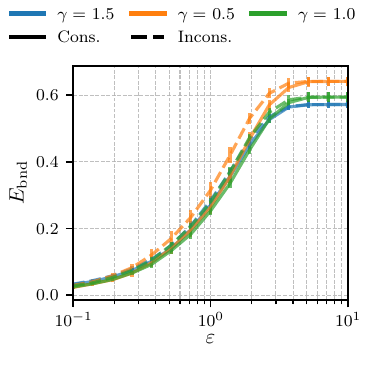}
    \caption{
        Comparison of consistent and inconsistent cases as a function of the attack strength.
        \textit{(Left, Center)}
        Case of the well-specified model presented in~\Cref{sec:wellspe}.
        We show both the probability of existence as per~\Cref{prop:exis} and the value of the boundary error.
        The error bars refer to 10 repetitions of the experiments for $d=500$.
        The metrics consider the $\what$ trained with $\alpha = 1.0$, $\lambda = 10^{-2}$, $r = 0.0$ and $\defnorm = 2$.
        \textit{(Right)}
        Case of the latent space model presented in~\Cref{sec:latent-space-model}.
        The error bars refer to 10 repetitions of the experiments for $d=512$.
        The metrics consider the $\what$ trained with $\alpha = 1.0$, $\lambda = 10^{-2}$, $r = 0.0$ and $\defnorm = 2$.
    }
    \label{fig:app:comparison-boundary-con-inconst}
\end{figure}

\begin{figure}
    \centering
    \hfill
    \includegraphics[width=0.28\textwidth]{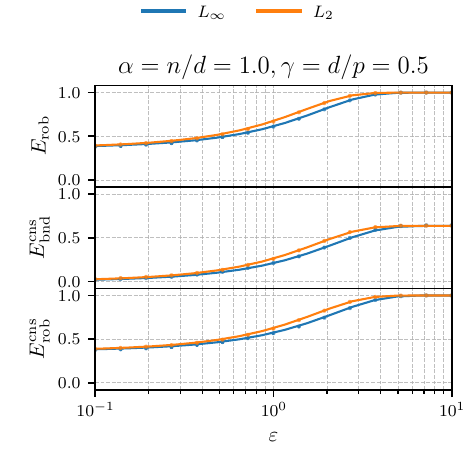}
    \hfill
    \includegraphics[width=0.28\textwidth]{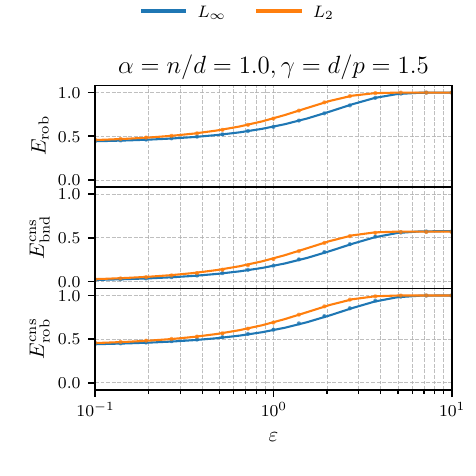}
    \caption{
        Behavior of the different adversarial errors for the Latent Space Model of~\Cref{sec:latent-space-model}. Here we show the behavior of the errors considered in the main text for different choices of $\gamma = \sfrac{d}{p}$ with $\gamma = 0.5$ for \textit{(Left)} and $\gamma = 1.5$ for \textit{(Right)}. We see that qualitatively the behaviors are the same in the two regimes.
    }
    \label{fig:app:comparison-attacks-latent-space}
\end{figure}

\begin{figure}[b]
    \centering
    \includegraphics[width=0.48\linewidth]{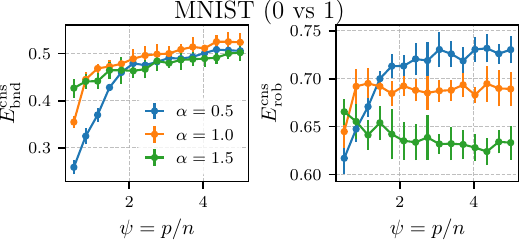}
    \hfill
    \includegraphics[width=0.48\linewidth]{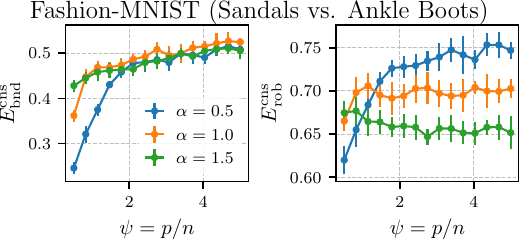}
    \caption{
    Empirical validation of the latent variable model predictions on real image datasets using complex Random Fourier Features.
    \textit{(Left)} Results on MNIST binary classification (digits 0 vs 1) with $d = 784$ features.
    \textit{(Right)} Results on Fashion-MNIST binary classification (Sandals vs. Ankle Boots) with $d = 784$ features.
    Points represent empirical averages over $10$ repetitions with error bars indicating standard deviation.
    The qualitative behavior matches the one seen for the Gaussian model in~\Cref{fig:latent-space-sweeps}.
    }
    \label{fig:app:mnist-fashion-mnist-experiments}
\end{figure}

We explore the behavior of the error metrics defined
in~\cref{eq:def-boundary-proper-latent-space,eq:def-misclass-proper-latent-space}
in~\Cref{fig:app:dependance-overparam-robust-proper,fig:app:dependance-overparam-boundary-proper}.
We see that the behavior is similar to the one of the model defined in the main text.
\revision{The black line in the same figure shows the performance of the well-specified model from the main text.}
Crucially we see also in this case the metric $\adverrproper$ equals the value of the clean generalization error in the $\varepsilon \to 0^+$ limit and it reaches one in the $\varepsilon \to \infty$ limit.
We have that $\bounderrproper$ is zero in the $\varepsilon \to 0^+$ limit.

\paragraph{Errors as a function of attack strength}
We additionally compare the consistent and inconsistent formulation of the boundary error as a function of the attack strength in \Cref{fig:app:comparison-boundary-con-inconst}.
The consistent boundary error is defined in~\cref{eq:boundary-proper} while the inconsistent one is defined as
\begin{equation}
    E_{\mathrm{bnd}} =
    \mathbb{E}_{(\x,y)}\qty[
        \max_{
            \substack{
                \attack\in B_{\attnorm}(\varepsilon)
            }
        }
        \mathbbm{1}\qty{
        \genfrac{}{}{0pt}{}{
            y
            \neq
            \hat{y}(\x + \attack)
        }{
            y = \hat{y}(\x)
        }
        }
    ] .
\end{equation}
\revision{where the crucial difference is the removal of the target-score invariance condition $g_\star(\x) = g_\star(\x + \attack)$.}
We have that also in this case consistent attacks produce a milder increase in the boundary error but the qualitative behavior is the same.

In~\Cref{fig:app:comparison-attacks-latent-space} we show the behavior of the different attacks considered in the main text for the latent space model. The qualitative behavior of the different models (different choice of $\gamma$) is the same.
\revision{As in the well-specified model, under the scaling of $\varepsilon$ in~\Cref{asm:scaling-eps-well-spec}, the $\ell_2$-bounded attack is more effective at increasing the error than the $\ell_\infty$-attack.}

\paragraph{Experiments on MNIST \citep{deng2012mnist} and Fashion-MNIST \citep{xiao2017fashion} datasets}
In~\Cref{fig:app:mnist-fashion-mnist-experiments} we show the empirical results on MNIST binary classification (digits 0 vs 1) and the Fashion-MNIST dataset for binary classification (Sandals vs Ankle Boots) using a teacher-student random features model.
We load MNIST images, center them by subtracting the dataset mean, and denote the centered images as $\tilde{\x}_\mu \in \mathbb{R}^{784}$.
In the complex Random Fourier Features (RFF) setting, we generate a random Gaussian projection matrix $\Omega \in \mathbb{R}^{D_{\mathrm{rff}} \times 784}$ with $D_{\mathrm{rff}} = 392$ and construct latent features $\z_\mu = [\text{Re}(\exp(-i \Omega \tilde{\x}_\mu));\, \text{Im}(\exp(-i \Omega \tilde{\x}_\mu))] \in \mathbb{R}^{2 D_{\mathrm{rff}}}$, so that the latent dimension is $d = 2 D_{\mathrm{rff}} = 784$.
A teacher classifier $\w_\star \in \mathbb{R}^{d}$ is trained via LinearSVC (no intercept) \citep{scikit-learn} on these latent features, and labels are generated through a noiseless channel.
The student observes transformed features $\x_\mu = \featmat \z_\mu + \uu_\mu$ as per~\Cref{asm:data-distr-latent-space}.
We train the student via $\ell^{\infty}$ ($s=\infty$) adversarial training~\cref{eq:risk-definition,eq:risk-minimization} with ridge regularization $\lambda = 10^{-2}$ and training attack budget $r = 1.0$.
We evaluate the consistent adversarial metrics defined in~\Cref{def:metrics}: consistent robust error $E^{\text{cns}}_{\text{rob}}$, and consistent boundary error $E^{\text{cns}}_{\text{bnd}}$ using $n_{\text{gen}} = 1000$ test samples with attack budget $\varepsilon = 1.0$.
Results are averaged over $10$ random seed repetitions with $d = 784$ fixed, varying the sample complexity $\alpha = n/d$ and overparameterization level $\gamma = d/p$.

\revision{The results are presented in~\Cref{fig:app:mnist-fashion-mnist-experiments}. The qualitative behavior is consistent with the Gaussian theory in~\Cref{sec:latent-space-model,sec:latent-space-overparam}. The consistent boundary error $\bounderrproper$ increases with overparameterization, while the consistent robust error $\adverrproper$ increases or decreases depending on the number of data points (the value of $\alpha = \sfrac{n}{d}$).}

\subsection{Neural network experiment}\label{sec:app:nn-experiment}
To test whether the dual role of overparameterization predicted by our theory (\Cref{prop:extreme-overparam}) persists beyond the linear estimator, we repeat the analysis replacing the linear model by a nonlinear neural network while keeping the data-generating process identical to~\Cref{asm:data-distr-latent-space}. This isolates the effect of the model class: the latent ground truth $\wstar$, the labels, and the attack geometry are unchanged, and only the estimator differs.

The estimator is a two-layer ReLU multilayer perceptron with hidden dimension $256$, trained to minimize the binary cross-entropy loss with Adam (learning rate $10^{-3}$, weight decay $10^{-4}$) and early stopping (patience $50$ on a $20\%$ validation split). We fix the latent dimension $d=1000$ and vary the feature dimension $p$ from $1000$ to $16000$, so that the overparameterization ratio $\psi=\sfrac{p}{n}$ spans more than one order of magnitude.
Adversarial perturbations are computed in the latent space with $\ell_{2}$-PGD ($50$ steps, $5$ random restarts, $\varepsilon=0.05$), enforcing the consistency constraint $\langle \wstar,\attack\rangle=0$ by projection at each step. All metrics are evaluated on a held-out test set and averaged over $10$ independent seeds.

\begin{figure}[t]
\centering
\includegraphics[width=0.65\linewidth]{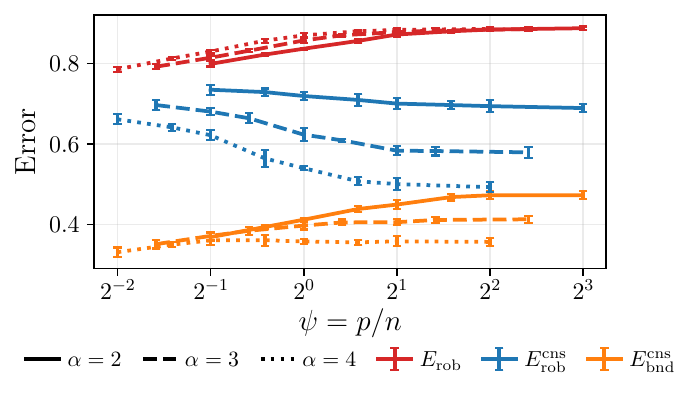}
\caption{Consistent metrics for a two-layer ReLU MLP trained on the latent variable model of~\Cref{asm:data-distr-latent-space}, as a function of the overparameterization $\psi=\sfrac{p}{n}$ ($d=1000$, $\ell_{2}$ attack with $\varepsilon=0.05$), for three sample complexities $\alpha=\sfrac{n}{d}\in\{2,3,4\}$. Replacing the linear estimator by a nonlinear network preserves the dual behavior predicted by the theory: $\advgenerr$ and $\bounderrproper$ increase with $\psi$, while $\adverrproper$ decreases. Averages over $10$ seeds; error bars are one standard deviation.}
\label{fig:nn-experiment}
\end{figure}

\Cref{fig:nn-experiment} reports the three metrics as a function of $p$. The qualitative predictions of the linear theory carry over to the nonlinear estimator: as overparameterization increases, the standard robust error $\advgenerr$ and the consistent boundary error $\bounderrproper$ grow, while the overall consistent robust error $\adverrproper$ decreases, driven by the improvement in clean generalization. This confirms that the dual effect of overparameterization is not an artifact of the linear model class.

\section{Proofs of the Results in the Main Text}\label{sec:app:proof-main}

The proofs in this appendix are organized according to the structure of the main results. We first present the finite-dimensional proofs in \Cref{sec:app:proof-finite-d}, next, in \Cref{sec:app:proof-high-d}, we provide proofs for the high-dimensional asymptotic results, including the derivation of the self-consistent equations stated in \Cref{thm:self-consistent-equations-latent-space}.
\revision{Finally, in \Cref{sec:app:limiting-error-metrics} we detail the derivation of the limiting error metrics for both the well-specified and latent-space models.}

\subsection{Proofs for the finite-dimensional results}\label{sec:app:proof-finite-d}

In this section we gather the proofs of the finite-dimensional results presented in the main text.

\begin{proof}[Proof of \Cref{prop:exis}]
    As discussed above, a consistent attack must satisfy the three conditions in \cref{eq:admissible,eq:modelflip}. First, note that this is only possible if $\hat{\w}_{\perp} \neq 0$, otherwise any admissible perturbation would \emph{a fortiori} violate \cref{eq:modelflip}. Therefore, from now on we assume $\hat{\w}_{\perp} \neq 0$. Consider an admissible attack $\attack\in H_{\attnorm}(\varepsilon)$. Since $-\attack \in H_{\attnorm}(\varepsilon)$, we can always fix the sign of $\attack$ to satisfy the constraint $\sign(\langle \hat{\w},\attack\rangle) = -\sign(\langle \hat{\w},\x\rangle)$. The restrictive condition is the margin $|\langle \hat{\w},\attack\rangle| > |\langle \hat{\w},\x\rangle|$. Since $\langle \attack,\wstar\rangle =0$, we have $\langle \hat{\w},\attack\rangle = \langle \hat{\w}_{\perp},\attack\rangle$, and hence the margin condition is equivalent to $|\langle \hat{\w}_{\perp},\attack\rangle| > |\langle \hat{\w},\x\rangle|$. This condition is satisfied if and only if it is satisfied by the supremum:
    \begin{align}
        \underset{\attack \in H_{\attnorm}(\varepsilon)}{\sup} |\langle \hat{\w}_{\perp},\attack\rangle| > |\langle \hat{\w},\x\rangle|\,.
    \end{align}
    Standard results from convex analysis imply that this supremum equals $\varepsilon$ times the $\ell^{q^{\star}}$-distance from $\hat{\w}_{\perp}$ to ${\rm span}(\wstar)$ (the metric projector):
    \begin{align}
    \label{eq:proof:dist}
       \underset{\attack \in H_{\attnorm}(\varepsilon)}{\sup} |\langle \hat{\w}_{\perp},\attack\rangle| = \varepsilon \,\inf_{\mu\in\mathbb{R}}\norm{\hat{\w}_{\perp}-\mu\wstar}_{q^{\star}} \eqqcolon \varepsilon\, d^{\star}_{q^{\star}}(\hat{\w}_{\perp})\,,
    \end{align}
    where $q^{\star}$ is the H\"older conjugate of $\attnorm$: $\sfrac{1}{\attnorm}+\sfrac{1}{q^{\star}}=1$.
\end{proof}

\begin{proof}[Proof of \Cref{cor:incon}]
    The existence part follows the same proof as in~\Cref{prop:exis}, but without the orthogonality constraint. We then have:
    \begin{align}
        \underset{\attack \in B_{\attnorm}(\varepsilon)}{\sup} |\langle \hat{\w},\attack\rangle| = \varepsilon||\hat{\w}||_{q^{\star}}.
    \end{align}
    \revision{Finally, $H_{\attnorm}(\varepsilon)\subseteq B_{\attnorm}(\varepsilon)$,
    so pointwise maximization over the two feasible sets gives
    $\adverrproper\leq\advgenerr$.}
\end{proof}

\begin{proof}[Proof of~\Cref{corr:well:Gauss:general}]
Conditionally on $\hat{\w}$, we have $\langle \hat{\w},\x\rangle \sim \mathcal{N}(0, \hat{\w}^{\top}\boldsymbol{\Sigma}\hat{\w})$, so $\langle \hat{\w},\x\rangle \overset{d}{=} \sqrt{\hat{\w}^{\top}\boldsymbol{\Sigma}\hat{\w}} \cdot Z$ with $Z\sim\mathcal{N}(0,1)$. The condition $\varepsilon d^{\star}_{q^{\star}}(\hat{\w}_{\perp}) \geq |\langle \hat{\w},\x\rangle|$ from~\Cref{eq:cond-well-spec} yields $\mathbb{P}[|Z| \leq \varepsilon d^{\star}_{q^{\star}}(\hat{\w}_{\perp})/\sqrt{\hat{\w}^{\top}\boldsymbol{\Sigma}\hat{\w}}] = 2\Phi(\varepsilon d^{\star}_{q^{\star}}(\hat{\w}_{\perp})/\sqrt{\hat{\w}^{\top}\boldsymbol{\Sigma}\hat{\w}}) - 1$.
{
We next prove the error comparison, still at fixed dimension and conditionally
on $\hat{\w}$. Write $b=\|\hat{\w}\|_{q^{\star}}$,
$a=d^{\star}_{q^{\star}}(\hat{\w}_{\perp})$, and $\rho=a/b\in[0,1]$.
Let $Z_1$ and $Z_2$ be the standardized target and model fields,
respectively, with correlation $c\in[-1,1]$. Under the noiseless link the
standardized signed margin is $M=\sign(Z_1)Z_2$. If $F_c$ denotes its c.d.f.,
the support functions of $H_{\attnorm}(\varepsilon)$ and
$B_{\attnorm}(\varepsilon)$ give
\begin{equation}\label{eq:app:finite-gaussian-error-rescaling}
\adverrproper(\varepsilon)
=F_c\!\left(\frac{\varepsilon a}{\sqrt{\hat{\w}^{\top}\boldsymbol{\Sigma}\hat{\w}}}\right)
=\advgenerr(\rho\varepsilon),
\qquad
\advgenerr(\varepsilon)
=F_c\!\left(\frac{\varepsilon b}{\sqrt{\hat{\w}^{\top}\boldsymbol{\Sigma}\hat{\w}}}\right).
\end{equation}
It remains to compare the c.d.f. at the two radii. Let $\phi$ and $\Phi$
denote the standard normal density and c.d.f. For $t\geq0$,
\begin{equation*}
F_c(t)=2\,\mathbb P(Z_1>0,Z_2\leq t),
\qquad
f_c(t)=F_c'(t)
=2\phi(t)\Phi\!\left(\frac{ct}{\sqrt{1-c^2}}\right)
\leq2\phi(t),
\end{equation*}
where the cases $c=\pm1$ follow by continuity. If $\Phi_2(\cdot,\cdot;c)$
is the standard bivariate Gaussian c.d.f. with correlation $c$, then
$F_c(t)=2[\Phi(t)-\Phi_2(0,t;c)]$. Since
$\partial_c\Phi_2(0,t;c)$ is the positive bivariate Gaussian density at
$(0,t)$, $F_c(t)$ is non-increasing in $c$. Hence
$F_c(t)\geq F_1(t)=2\Phi(t)-1$. Consequently,
\begin{equation*}
F_c(t)-t f_c(t)
\geq 2\Phi(t)-1-2t\phi(t)\geq0,
\end{equation*}
because the last expression vanishes at $t=0$ and has derivative
$2t^2\phi(t)\geq0$. Thus $t\mapsto F_c(t)/t$ is non-increasing on
$(0,\infty)$, and $F_c(\rho t)\geq\rho F_c(t)$. Combining this with
\eqref{eq:app:finite-gaussian-error-rescaling} and monotonicity of $F_c$
proves
$\rho\advgenerr(\varepsilon)\leq\adverrproper(\varepsilon)
\leq\advgenerr(\varepsilon)$. The case $\varepsilon=0$ follows directly.
}
\end{proof}

\begin{proof}[Proof of~\Cref{rmk:powerlaw}]
Assume $q=2$ so that $q^\star=2$ and $d^\star_{q^\star}(\cdot)=\|\cdot\|_2$.
Consider the power-law spectrum
\[
\lambda_i \;=\; \frac{d\, i^{-\beta}}{Z_\beta(d)},
\qquad
Z_\beta(d)=\sum_{j=1}^d j^{-\beta},
\qquad (\beta>0),
\]
with orthonormal eigenbasis $\{v_i\}_{i=1}^d$, and let $\wstar=\sqrt d\,v_1$.
Fix a correlation level $m\in[0,1]$ and consider predictors of the form
\[
\hat \w \;=\; m\sqrt d\,\vv_1 + \hat \w_\perp,
\qquad
\|\hat \w_\perp\|_2^2 \;=\; d(1-m^2).
\]
Write $\hat \w_\perp=\sum_{i=2}^d c_i v_i$ with $\sum_{i=2}^d c_i^2=d(1-m^2)$.
The Gaussian logit appearing in~\eqref{eq:probGauss-well-spec-general} equals
\[
\gamma(\hat \w)
\;:=\;
\frac{\varepsilon\|\hat \w_\perp\|_2}{\sqrt{\hat \w^\top \boldsymbol{\Sigma} \hat \w}}
\;=\;
\frac{\varepsilon\sqrt{d(1-m^2)}}{\sqrt{\hat \w^\top \boldsymbol{\Sigma} \hat \w}}.
\]
Thus, for $q=2$, comparing predictors reduces to comparing the denominators
$\sqrt{\hat \w^\top \boldsymbol{\Sigma} \hat \w}$.

Using $\boldsymbol{\Sigma} \vv_i=\lambda_i \vv_i$ and orthonormality,
\begin{align}
\hat \w^\top \boldsymbol{\Sigma} \hat \w =
m^2 d\, \lambda_1 + \sum_{i=2}^d c_i^2 \lambda_i =
\frac{d^2}{Z_\beta(d)}\Big(m^2 + (1-m^2)\sum_{i=2}^d p_i\, i^{-\beta}\Big),
\label{eq:quad-form-pl}
\end{align}
where we introduced the weights
\[
p_i \;:=\; \frac{c_i^2}{d(1-m^2)} \ge 0,
\qquad \sum_{i=2}^d p_i = 1.
\]

Fix $k\in\{1,\dots,d-2\}$ and define two predictors with the same $m$ and the same
$\|\hat \w_\perp\|_2^2=d(1-m^2)$ but different allocation of $\hat \w_\perp$:
\[
\hat \w_\perp^{\mathrm{LR}}
:= \sqrt{\frac{d(1-m^2)}{k}}\sum_{i=2}^{k+1} \vv_i,
\qquad
\hat \w_\perp^{\mathrm{Tail}}
:= \sqrt{\frac{d(1-m^2)}{d-k-1}}\sum_{i=k+2}^{d} \vv_i.
\]
Equivalently, $p_i$ is uniform on $S_{\mathrm{LR}}=\{2,\dots,k+1\}$ (resp. on
$S_{\mathrm{Tail}}=\{k+2,\dots,d\}$). Define the corresponding averages
\[
\mu_{\mathrm{LR}}
:=\frac1k\sum_{i=2}^{k+1} i^{-\beta},
\qquad
\mu_{\mathrm{Tail}}
:=\frac1{d-k-1}\sum_{i=k+2}^{d} i^{-\beta}.
\]
Then~\eqref{eq:quad-form-pl} yields
\[
\hat \w^\top\boldsymbol{\Sigma} \hat \w\Big|_{\mathrm{LR}}
=
\frac{d^2}{Z_\beta(d)}\Big(m^2+(1-m^2)\mu_{\mathrm{LR}}\Big),
\qquad
\hat \w^\top\boldsymbol{\Sigma} \hat \w\Big|_{\mathrm{Tail}}
=
\frac{d^2}{Z_\beta(d)}\Big(m^2+(1-m^2)\mu_{\mathrm{Tail}}\Big).
\]

Since $\beta>0$, the sequence $a_i:=i^{-\beta}$ is strictly decreasing.
For every $(i,j)\in S_{\mathrm{LR}}\times S_{\mathrm{Tail}}$ we have $i<j$, hence
$a_i>a_j$. Summing over all $k(d-k-1)$ such pairs gives
\[
(d-k-1)\sum_{i=2}^{k+1} i^{-\beta}
\;>\;
k\sum_{j=k+2}^{d} j^{-\beta}.
\]
Dividing by $k(d-k-1)$ yields $\mu_{\mathrm{LR}}>\mu_{\mathrm{Tail}}$.

Because $m^2+(1-m^2)\mu$ is strictly increasing in $\mu$, the inequality
$\mu_{\mathrm{LR}}>\mu_{\mathrm{Tail}}$ implies
\[
\hat \w^\top\boldsymbol{\Sigma} \hat \w\Big|_{\mathrm{LR}}
\;>\;
\hat \w^\top\boldsymbol{\Sigma} \hat \w\Big|_{\mathrm{Tail}}
\quad\Longrightarrow\quad
\gamma_{\mathrm{Tail}}>\gamma_{\mathrm{LR}}.
\]
Moreover, the multiplicative separation between logits is
\[
\frac{\gamma_{\mathrm{Tail}}}{\gamma_{\mathrm{LR}}}
=
\sqrt{
\frac{m^2+(1-m^2)\mu_{\mathrm{LR}}}
     {m^2+(1-m^2)\mu_{\mathrm{Tail}}}
}
=
\sqrt{
\frac{m^2+(1-m^2)\frac1k\sum_{i=2}^{k+1} i^{-\beta}}
     {m^2+(1-m^2)\frac1{d-k-1}\sum_{i=k+2}^{d} i^{-\beta}}
}
\;>\;1.
\]
Since $\Phi$ is increasing, this implies
$2\Phi(\gamma_{\mathrm{Tail}})-1 > 2\Phi(\gamma_{\mathrm{LR}})-1$,
i.e.\ tail-aligned predictors are more vulnerable than LR-aligned predictors
by the explicit constant factor above (at the logit level).
\end{proof}

The proofs of~\Cref{prop:exist:latent,corr:prob-latent} follow the same lines as the proofs above.

\subsection{Proof of the high-dimensional asymptotic results}\label{sec:app:proof-high-d}

In this section, we provide rigorous proofs for the theoretical results presented in the main paper, focusing on \Cref{thm:self-consistent-equations-latent-space}.

Central to our analysis is the Convex Gaussian MinMax Theorem (CGMT), a fundamental tool that bridges complex high-dimensional optimization problems with simpler low-dimensional counterparts. The CGMT enables us to transform our challenging primary optimization problem into a more tractable auxiliary problem, ultimately leading to the self-consistent equations presented in \Cref{eq:latent-space-eq-channel,eq:latent-space-eq-prior-gamma-big,eq:latent-space-eq-prior-gamma-small}.

We begin by stating the CGMT in its general form.

\begin{theorem}[CGMT \cite{gordon_1988,thrampoulidis2015gaussian}]\label{thm:cgmt}
Let \(\boldsymbol{G} \in \RR^{m \times n}\) be an i.i.d. standard normal matrix and \(\mathbf{g} \in \RR^m\), \(\mathbf{h} \in \RR^n\) two i.i.d. standard normal vectors independent of each other. For compact sets \(\mathcal{S}_{\w} \subset \RR^n\) and \(\revision{\mathcal{S}_{\boldsymbol{u}} \subset \RR^m}\), consider the following optimization problems with continuous function \(\psi\) on \(\mathcal{S}_{\w} \times \mathcal{S}_{\boldsymbol{u}}\):
\begin{align}
    \mathbf{C}(\boldsymbol{G}) &:= \min_{\w \in \mathcal{S}_{\w}} \max _{\boldsymbol{u} \in \mathcal{S}_{\boldsymbol{u}}}
    \boldsymbol{u}^{\top} \boldsymbol{G} \w +
    \psi(\w, \boldsymbol{u}) \\
    \mathcal{C}(\mathbf{g}, \mathbf{h}) &:= \min_{\w \in \mathcal{S}_{\w}} \max _{\boldsymbol{u} \in \mathcal{S}_{\boldsymbol{u}}}
    \|\w\|_2 \mathbf{g}^{\top} \boldsymbol{u} +
    \|\boldsymbol{u}\|_2 \mathbf{h}^{\top} \w +
    \psi(\w, \boldsymbol{u})
\end{align}
The following statements hold:
\begin{enumerate}
    \item For all \(c\in \RR\): \(\mathbb{P}(\mathbf{C}(\boldsymbol{G})<c) \le 2 \mathbb{P}(\mathcal{C}(\mathbf{g}, \mathbf{h}) \le c)\)
    \item When \(\mathcal{S}_{\w}\) and \(\mathcal{S}_{\boldsymbol{u}}\) are convex sets and \(\psi\) is convex-concave on \(\mathcal{S}_{\w} \times \mathcal{S}_{\boldsymbol{u}}\), for all \(c\in \RR\): \(\mathbb{P}(\mathbf{C}(\boldsymbol{G})>c) \le 2 \mathbb{P}(\mathcal{C}(\mathbf{g}, \mathbf{h}) \ge c)\)
    \item Consequently, for all \(\mu \in \RR\), \(t > 0\): \(\mathbb{P}(|\mathbf{C}(\boldsymbol{G})-\mu|>t) \le 2 \mathbb{P}(|\mathcal{C}(\mathbf{g}, \mathbf{h})-\mu| \ge t)\)
\end{enumerate}
\end{theorem}

We will utilize a specialized version of the CGMT developed by \cite{loureiro2021learning} for generalized linear models.

\subsubsection{Mathematical Preliminaries}

Our analysis relies heavily on Moreau envelopes and proximal operators from convex analysis. These concepts have become essential tools in the asymptotic analysis of high-dimensional convex problems \citep{boyd_vandenberghe_2004,parikh_boyd_proximal_algo}. We provide key definitions below.

\begin{definition}[Moreau Envelope]\label{def:app:moreau-definition}
For a convex function $f : \RR^n \to \RR$, its Moreau envelope is defined as:
\begin{equation}\label{eq:app:moreau-definition}
    \mathcal{M}_{V f(\cdot)} (\boldsymbol{\omega}) = \min_{\x} \qty[\frac{1}{2 V} \norm{\x - \boldsymbol{\omega}}_2^2 + f(\x)]
\end{equation}
where $\mathcal{M}_{V f(\cdot)} : \RR^n \to \RR$.
\end{definition}

\begin{definition}[Proximal Operator]\label{def:app:proximal-definition}
For a convex function $f : \RR^n \to \RR$, its Proximal operator is defined as:
\begin{equation}\label{eq:app:proximal-definition}
    \mathcal{P}_{V f(\cdot)} (\boldsymbol{\omega}) = \argmin_{\x} \qty[\frac{1}{2 V}\norm{\x - \boldsymbol{\omega}}_2^2 + f(\x)]
\end{equation}
where $\mathcal{P}_{V f(\cdot)} : \RR^n \to \RR^n$.
\end{definition}

\begin{theorem}[Gradient of Moreau Envelope \citep{thrampoulidis2018precise}, Lemma D1]\label{thm:app:envelope-theorem}
For a convex function $f : \RR^n \to \RR$ with Moreau envelope $\mathcal{M}_{V f(\cdot)}$ and Proximal operator $\mathcal{P}_{V f(\cdot)}$:
\begin{equation}\label{eq:app:envelope-thm}
    \grad_{\boldsymbol{\omega}} \mathcal{M}_{V f(\cdot)}(\boldsymbol{\omega})= \frac{1}{V} \left(\boldsymbol{\omega} - \mathcal{P}_{V f(\cdot)}(\boldsymbol{\omega})\right)
\end{equation}
\end{theorem}

Additionally, we will use these important properties:
\begin{equation}\label{eq:app:proximal-moreau-shift}
    \mathcal{M}_{V f(\cdot + \boldsymbol{u})} (\boldsymbol{\omega}) = \mathcal{M}_{V f(\cdot)} (\boldsymbol{\omega} + \boldsymbol{u}) \,, \quad
    \mathcal{P}_{V f(\cdot + \boldsymbol{u})} (\boldsymbol{\omega}) = \revision{\mathcal{P}_{V f(\cdot)} (\boldsymbol{\omega} + \boldsymbol{u})-\boldsymbol{u}}
\end{equation}
which follow directly from a change of variables in the minimization.

\begin{definition}[Dual of a number]\label{def:app:dual-number}
For $a \in (1,\infty]$, we define its dual $a^\star \in [1,\infty)$ as the unique number satisfying $\sfrac{1}{a} + \sfrac{1}{a^\star} = 1$, with the conventions $1^\star = \infty$ and $\infty^\star = 1$.
\end{definition}

\subsubsection{Assumptions and Preliminary Discussion}

{
The main text uses normalized covariates, $\z\sim\mathcal N(0,\idmat_d/d)$ and $\uu\sim\mathcal N(0,\idmat_p/p)$, with unscaled prediction fields. To match the standard CGMT normalization, in this subsection only we set $\tilde\z=\sqrt d\,\z$, $\tilde\uu=\sqrt p\,\uu$, $\tilde\x=\sqrt p\,\x$, and $\tilde\featmat=\sqrt{p/d}\,\featmat$, and then drop the tildes. Thus the fields become $\wstar^\top\z/\sqrt d$ and $\w^\top\x/\sqrt p$. This exact change of variables leaves the order parameters in~\Cref{thm:self-consistent-equations-latent-space} unchanged.

\begin{assumption}[Estimation from the dataset]\label{ass:app:estimation-from-dataset}
Given a dataset $\mathcal{D}$ made of $n$ input-output pairs $\{(\x_i, y_i)\}_{i=1}^{n}$, where $\x_i \in \RR^p$ and $y_i \in\RR$, we estimate
\begin{equation}
    \what \in \argmin_{\w \in \RR^p} \sum_{\dataidx = 1}^{\nsamples}
    \max_{
        \norm{\defend_\dataidx}_{\defnorm} \leq r p^{\sfrac{1}{2}-\sfrac{1}{\defnorm^\star}}
    }
    \lossfun \qty(y_\dataidx \frac{\w^\top \qty(\x_\dataidx + \defend_\dataidx)}{\sqrt{p}})
    + \lambda \norm{\w}_2^2 \,,
\end{equation}
where $\lossfun : \RR \to \RR$ is a convex non-increasing function and where the second term is a convex regularization function whose strength can be tuned with $\lambda \in [0, \infty)$.
\end{assumption}

\begin{assumption}[Data Distribution]\label{ass:app:data-distribution}
In this rescaled convention, $\z \sim \mathcal{N}(\mathbf{0},\idmat_d)$, $f_{\wstar}(\z) = \varphi(\langle \wstar,\z\rangle/\sqrt d)$ with $\wstar \in\mathbb{S}^{d-1}(\sqrt{d})$, and $\x = \featmat \z + \uu\in\mathbb R^p$ with independent $\uu \sim \mathcal{N}(\mathbf 0,\idmat_p)$ and
\begin{equation}\label{eq:app:feature-mat-definition}
    \featmat = \begin{cases}{
        \left[\begin{array}{c}
        \sqrt{\frac{p}{d}} \idmat_d \\
        \mathbf{0}_{(p-d) \times d}
        \end{array}\right]} & \text { if } p \geq d \\
        {\left[\begin{array}{ll}
            \idmat_p & \mathbf{0}_{p \times(d-p)}
        \end{array}\right]} & \text { if } p<d
    \end{cases}\,.
\end{equation}\end{assumption}

\begin{assumption}[High-Dimensional Limit]\label{ass:app:high-dimensional-limit}
We consider the proportional high-dimensional regime where both the number of training data and input dimension $n, d, p \to \infty$ at a fixed ratio $\alpha := \sfrac{n}{d}$ and $\psi = \sfrac{p}{n}$.
\end{assumption}

This setting considers most of the losses used in machine learning setups for binary classification, \textit{e.g.} logistic, hinge, exponential losses.
We additionally remark that with the given choice of regularization the whole cost function is coercive.

\begin{assumption}[Scaling of Adversarial Norm Constraint]\label{ass:app:scaling-eps}
For observed dimension $p$ and training norm $s>1$, the dimension-free budget $r$ in the main text corresponds here to physical radius $r p^{1/2-1/s^\star}$. Hence $p^{-1/2}\sup_{\|\defend\|_s\leq r p^{1/2-1/s^\star}}\w^\top\defend=r(\|\w\|_{s^\star}^{s^\star}/p)^{1/s^\star}=O(1)$.
\end{assumption}

\subsubsection{Problem Simplification}

Recall that we start from the following optimization problem:
\begin{equation}\label{eq:app:adversarial-training-problem}
    \Phi_p = \min_{\w \in \RR^p} \sum_{\dataidx = 1}^{n}
    \max_{
        \norm{\defend_\dataidx}_{\defnorm} \leq r p^{\sfrac{1}{2}-\sfrac{1}{\defnorm^\star}}
    }
    \lossfun \qty(y_\dataidx \frac{\w^\top \qty(\x_\dataidx + \defend_\dataidx)}{\sqrt{p}})
    + \lambda \norm{\w}_2^2 \,.
\end{equation}
The non-increasing property of \(\lossfun\) allows us to simplify the inner maximization, leading to an equivalent formulation
\begin{equation}
    \Phi_p = \min_{\w \in \RR^p} \sum_{\dataidx = 1}^{n}
    \lossfun \qty(
        y_\dataidx \frac{\w^\top \x_\dataidx}{\sqrt{p}}
        - \frac{r}{p^{1/\defnorm^\star}} \norm{\w}_{\defnorm^\star}
    )
    + \lambda \norm{\w}_2^2 \,.
\end{equation}

To facilitate our analysis, we introduce $P = \norm{\w}_{\defnorm^\star}^{\defnorm^\star} / p$ and its Lagrange multiplier $\hat{P}$, which decouple the norm constraint. This leads to
\begin{equation}
    \Phi_p = \min_{\w \in \RR^p, P} \max_{\hat{P}}
    \sum_{\dataidx = 1}^{\nsamples} \lossfun\qty(
        y_\dataidx \frac{\w^\top \x_\dataidx}{\sqrt{p}}
        - r \sqrt[\defnorm^\star]{P}
    )
    + \lambda \norm{\w}_2^2
    + \hat{P} \norm{\w}_{\defnorm^\star}^{\defnorm^\star} - p P \hat{P},
\end{equation}
This reformulation exposes the dimension-free adversarial shift $rP^{1/s^\star}$ and allows us to apply the CGMT.
}

\subsubsection{Scalarization and Application of CGMT}

To facilitate our analysis, we further introduce effective regularization and loss functions, \(\tilde{\regfun}\) and \(\tilde{\lossfun}\), respectively.
These functions are defined as
\begin{equation}
    \tilde{\lossfun}(\y, \z) = \sum_{\dataidx = 1}^{\nsamples} \lossfun\qty(
        y_\dataidx \z_\dataidx
        - \revision{r \sqrt[\defnorm^\star]{P}}
    ) \,, \quad
    \tilde{\regfun}(\w) = \norm{\w}_2^2 + \hat{P} \norm{\w}_{\defnorm^\star}^{\defnorm^\star}
    \,.
\end{equation}
\revision{A crucial step in our analysis involves inverting the order of the min-max optimization. We justify this operation by considering minimization with respect to \(\w \in \RR^p\) at fixed \(\hat{P}\) and \(P\). The objective is convex in \(\w\), and the relevant saddle formulation has convex constraint sets.}
Under these conditions, we apply Sion's minimax theorem, which guarantees the existence of a saddle point and allows us to interchange the order of minimization and maximization without affecting the optimal value.

We additionally notice that the data distribution defined in~\Cref{ass:app:data-distribution} fits the framework of \cite{loureiro2021learning}. Specifically, it can be seen as the case treated in Section~3.1 with the non-linearity reducing to additive Gaussian noise.

This reformulation enables us to directly apply \citep[Lemma~11]{loureiro2021learning}, a careful application of \Cref{thm:cgmt} to scenarios involving non-separable convex regularization and loss functions. The result is a lower-dimensional equivalent of our original high-dimensional minimization problem that captures the limiting behavior of the optimum.

Consequently, our analysis now focuses on a low-dimensional functional, which takes the form
\begin{equation}\label{eq:app:exrtemisation-problem-gordon}
    \Tilde{\Phi} = \min_{P, m, \eta, \tau_1} \max_{\hat{P}, \kappa, \tau_2, \nu}
    \Bigg[
        \frac{\kappa \tau_1}{2}
        - \alpha \mathcal{L}_{\lossfun}
        - \frac{\eta}{2\tau_2} \qty(\nu^2 + \kappa^2)
        - \frac{\eta \tau_2}{2}
        - \mathcal{L}_{\regfun}
        + m \nu
        - P \hat{P}
    \Bigg]
\end{equation}
\revision{where we have restored the min--max order of the problem.}

In this expression, $\gaussvecone$ and $\gaussvectwo$ are independent Gaussian vectors with i.i.d. standard normal components. The terms $\mathcal{L}_{\lossfun}$ and $\mathcal{L}_{\regfun}$ represent the scaled averages of Moreau Envelopes (\cref{eq:app:moreau-definition})
\begin{align}
    \mathcal{L}_{\lossfun} &= \frac{1}{n} \mathbb{E}\qty[
        \mathcal{M}_{\frac{\tau_1}{\kappa} \tilde{\lossfun}(\y, \cdot)}\qty(
            m \s + \eta \gaussvectwo
        )
    ] \\
    \mathcal{L}_{\regfun} & = \frac{1}{d} \mathbb{E}\qty[
        \mathcal{M}_{\frac{\eta}{\tau_2} \tilde{\regfun}(\cdot)}\qty(\frac{\eta}{\tau_2}\qty(\kappa \gaussvecone + \nu \wstar))
    ]
\end{align}
\revision{The extremization problem in~\cref{eq:app:exrtemisation-problem-gordon} captures the leading-order behavior of the original optimization problem in~\cref{eq:app:adversarial-training-problem} as $n,d,p\to\infty$.}

\revision{This dimensional reduction replaces the original high-dimensional problem with a tractable, dimension-free set of scalar equations.}

\revision{At finite size, the optimization problem \(\Tilde{\Phi}\) still depends implicitly on $n,d$, and $p$.}
We introduce two variables
\begin{equation}\label{eq:app:limiting-distrib}
    \w_{\rm eq} = \mathcal{P}_{\frac{\eta^*}{\tau_2^*} \tilde{\regfun}\left(\cdot\right)}\left(\frac{\eta^*}{\tau_2^*}\left(\nu^* \mathbf{t}+\kappa^* \mathbf{g}\right)\right) \, , \quad
    \z_{\rm eq} = \mathcal{P}_{\frac{\tau_1^*}{\kappa^*} \tilde{\lossfun}(\cdot,\mathbf{y})}\left(m^* \mathbf{s}+\eta^* \mathbf{h}\right)
\end{equation}
where $(\eta^\star, \tau_2^\star, P^\star, \hat{P}^\star, \kappa^\star, \nu^\star, m^\star,\tau_1^\star)$ are the extremizer points of $\Tilde{\Phi}$.

Building upon \citep[Theorem~5]{loureiro2021learning}, we can establish a convergence result. Let \(\what\) be an optimal solution of the problem defined in \cref{eq:app:adversarial-training-problem}, and let \(\hat{\z} = \frac{1}{\sqrt{p}} \boldsymbol{X} \what\), where $\boldsymbol{X}\in\RR^{n\times p}$ stacks the data points by row. For any Lipschitz function \(\varphi_1 : \RR^p \to \RR\) and any separable, pseudo-Lipschitz function \(\varphi_2 : \RR^n \to \RR\), there exist constants \(\epsilon, C, c > 0\) such that
\begin{align}
    & \mathbb{P}\left(\left|\varphi_1\left(\frac{\hat{\mathbf{w}}}{\sqrt{p}}\right)-\mathbb{E}\left[\varphi_1\left(\frac{\w_{\rm eq}}{\sqrt{p}}\right)\right]\right| \geq \epsilon\right) \leq \frac{C}{\epsilon^2} e^{-c n \epsilon^4} \label{eq:concentration-w}\\
    & \mathbb{P}\left(\left|\varphi_2\left(\frac{\hat{\mathbf{z}}}{\sqrt{n}}\right)-\mathbb{E}\left[\varphi_2\left(\frac{\z_{\rm eq}}{\sqrt{n}}\right)\right]\right| \geq \epsilon\right) \leq \frac{C}{\epsilon^2} e^{-c n \epsilon^4} \label{eq:concentration-z}
\end{align}

It demonstrates that the limiting values of any function depending on \(\what\) and \(\hat{\z}\) can be computed by taking the expectation of the same function evaluated at \(\w_{\rm eq}\) or \(\z_{\rm eq}\), respectively. This convergence property allows us to translate results from our low-dimensional proxy problem back to the original high-dimensional setting with high probability.

\subsubsection{Derivation of Saddle Point equations}

\revision{We next show that extremizing over $m,\eta,\tau_1,P,\hat{P},\nu,\tau_2,\kappa$ yields the optimal value $\Tilde{\Phi}$ in~\cref{eq:app:exrtemisation-problem-gordon}. We derive the saddle-point equations directly and then show that their high-dimensional limits coincide with those in the main text.}

We obtain the first set of derivatives, which depend only on the loss function and the channel part, by taking the derivatives with respect to $m,\eta,\tau_1,P$:
\begin{equation}\label{eq:app:cgmt-eq-channel-part}
\begin{aligned}
    \pdv{m} &: \nu = \alpha \frac{\kappa}{n \tau_1} \mathbb{E}\left[\left(\frac{m}{\eta} \mathbf{h}-\mathbf{s}\right)^{\top} \mathcal{P}_{\frac{\tau_1}{\kappa} \tilde{\lossfun}(., \mathbf{y})}\left(m \mathbf{s}+\eta \mathbf{h}\right)\right] \\
    \pdv{\eta} &: \tau_2=\alpha \frac{\kappa}{\tau_1} \eta-\frac{\kappa \alpha}{\tau_1 n} \mathbb{E}\left[\mathbf{h}^{\top} \mathcal{P}_{\frac{\tau_1}{\kappa} \tilde{\lossfun}(\cdot, \mathbf{y})}\left(m \mathbf{s}+\eta \mathbf{h}\right)\right] \\
    \pdv{\tau_1} &: \frac{\tau_1^2}{2}=\frac{1}{2} \alpha \frac{1}{n} \mathbb{E}\left[\left\|m \mathbf{s}+\eta \mathbf{h}-\mathcal{P}_{\frac{\tau_1}{\kappa} \tilde{\lossfun}(\cdot, y)}\left(m \mathbf{s}+\eta \mathbf{h}\right)\right\|_2^2\right] \\
    \pdv{P} &: \hat{P} = \frac{\alpha}{n} \partial_P \mathbb{E}\qty[
        \mathcal{M}_{\frac{\tau_1}{\kappa} \tilde{\lossfun}(\y, \cdot)}\qty(
            m \s + \eta \gaussvectwo
        )
    ]
\end{aligned}
\end{equation}
By taking the derivatives with respect to the remaining variables $\kappa, \nu, \tau_2, \hat{P}$ we obtain a set of equations depending on regularization and prior over the teacher weights
\begin{equation}\label{eq:app:cgmt-eq-prior-part}
\begin{aligned}
    \pdv{\kappa} &: \tau_1=\frac{1}{d} \mathbb{E}\left[\mathbf{g}^{\top} \mathcal{P}_{\frac{\eta}{\tau_2} \tilde{\regfun}(\cdot)}\left(\frac{\eta}{\tau_2}\left(\nu \wstar+\kappa \mathbf{g}\right)\right)\right] \\
    \pdv{\nu} &: m =\frac{1}{d} \mathbb{E}\left[\wstar^{\top} \mathcal{P}_{\frac{\eta}{\tau_2} \tilde{\regfun}(\cdot)} \left(\frac{\eta}{\tau_2}\left(\nu \wstar+\kappa \mathbf{g}\right)\right)\right] \\
    \pdv{\tau_2} &: \frac{1}{2 d} \frac{\tau_2}{\eta} \mathbb{E}\left[\left\|\frac{\eta}{\tau_2}\left(\nu \wstar+\kappa \mathbf{g}\right)-\mathcal{P}_{\frac{\eta}{\tau_2} \tilde{\regfun}(\cdot)} \left(\frac{\eta}{\tau_2}\left(\nu  \wstar+\kappa \mathbf{g}\right)\right)\right\|_2^2\right] =\frac{\eta}{2 \tau_2}\left(\nu^2 + \kappa^2\right) - m \nu - \kappa \tau_1 + \frac{\eta \tau_2}{2} + \frac{\tau_2}{2 \eta} m^2 \\
    \pdv{\hat{P}} &: P = \frac{1}{d}\partial_{\hat{P}} \mathbb{E}\qty[
        \mathcal{M}_{\frac{\eta}{\tau_2} \tilde{\regfun}(\cdot)}\qty(\frac{\eta}{\tau_2}\qty(\kappa \gaussvecone + \nu \wstar))
    ]
\end{aligned}
\end{equation}
\revision{Rewriting these equations in the form of~\Cref{thm:self-consistent-equations-latent-space} follows \citep[Appendix~C.2]{loureiro2021learning}: it requires two changes of variables and an integration by parts.}

To perform this rewriting the first ingredient we need is the following change of variables
\begin{equation}\label{eq:app:gordon-replica-dict}
\begin{aligned}
    m &\leftarrow m \,, & \quad \tau &\leftarrow \eta^2 + m^2 \,, & \quad V &\leftarrow \frac{\tau_1}{\kappa} \,, & \quad P&\leftarrow P \,, \\
    \bar{V} &\leftarrow \frac{\tau_2}{\eta} \,, & \quad \bar{\tau} &\leftarrow \kappa^2 \,, & \quad \bar{m} &\leftarrow \nu \,, & \quad \bar{P}&\leftarrow \hat{P} \,.
\end{aligned}
\end{equation}
and the use of Isserlis' theorem \citep{isserlis1918formula} to simplify the expectations involving the Gaussian vectors $\gaussvecone$ and $\gaussvectwo$.

\subsubsection{Rewriting of the Saddle Point Equations}

We now rewrite the zero-gradient equations in a form more amenable to numerical solution. Recently, \citet{vilucchio2025asymptotics} showed the relationship of these equations with the so-called ``state evolution'' for the AMP algorithm. This rewriting brings a computational advantage, since their numerical stability is improved.

To simplify the equations we first notice that, due to the separability, all terms depending on the proximal of either $\tilde{\lossfun}$ or $\tilde{\regfun}$ simplify the $n$ or $d$ at the denominator.
This cancellation is crucial as it eliminates the explicit dependence on the problem dimension, allowing us to derive dimension-independent equations.

To obtain specifically the form implied in the main text we introduce
\begin{equation}
    \mathcal{Z}_0(y, \omega, V) = \int \frac{\dd x}{\sqrt{2 \pi V}} e^{-\frac{1}{2 V}(x-\omega)^2} \delta\left(y-f^0(x)\right) \,.
\end{equation}
The function $\mathcal{Z}_0$ can be interpreted as a partition function of the conditional distribution $\mathbb{P}_{\rm out}$ and contains all of the information about the label generating process. This brings the first set of equations to be
\begin{equation}
    \begin{cases}
        \bar{m} = \alpha \mathbb{E}_{\xi}\left[
            \int_{\RR} \dd{y} \partial_\omega \mathcal{Z}_0 (y, \sqrt{\frac{m^2}{\tau}} \xi, 1- \frac{m^2}{\tau}) f_\ell(\sqrt{\tau} \xi, P, r)
        \right] \\
        \bar{\tau} = \alpha \mathbb{E}_{\xi}\left[
            \int_{\RR} \dd{y} \mathcal{Z}_0 (y, \sqrt{\frac{m^2}{\tau}} \xi, 1- \frac{m^2}{\tau})f_\ell^2(\sqrt{\tau} \xi, P, r)
        \right] \\
        \bar{V} = -\alpha \mathbb{E}_{\xi}\left[
            \int_{\RR} \dd{y} \mathcal{Z}_0 (y, \sqrt{\frac{m^2}{\tau}} \xi, 1- \frac{m^2}{\tau})\partial_\omega f_\ell(\sqrt{\tau} \xi, P, r)
        \right] \\
        \bar{P} = r P^{1/s} \alpha \mathbb{E}_{\xi} \qty[
            \int_{\RR} \dd{y} \mathcal{Z}_0 (y, \sqrt{\frac{m^2}{\tau}} \xi, 1- \frac{m^2}{\tau}) y f_\ell (\sqrt{\tau} \xi, P, r)
        ]
    \end{cases}
\end{equation}
where $\xi \sim \mathcal{N}(0,1)$ and we defined for notational convenience the following
\begin{equation}
    f_\ell(\omega, P, r) = \frac{1}{V} \qty(\mathcal{P}_{V\ell(y \cdot - r \sqrt[s^\star]{P})} (\omega) - \omega) \,.
\end{equation}

Next, we simplify the remaining set of equations by introducing
\begin{equation}
    \mathcal{Z}_{w_\star}(\varrho, \Lambda) = \int \dd{w} e^{-\frac{1}{2}w^2} e^{-\frac{\Lambda}{2} w^2 + \varrho w},
\end{equation}
which, in turn, leads to the second set of equations
\begin{align}
    &\begin{cases}
        \bar{m} = \frac{1}{\sqrt{\gamma}} \mathbb{E}_{\xi}\left[
            \partial_\varrho \mathcal{Z}_{w_\star}
            f^1_w
        \right] \\
        \bar{\tau} = (1+\gamma) \mathbb{E}_{\xi}\left[
            \mathcal{Z}_{w_\star}
            (f^1_w)^2
        \right]
        + (1-\gamma)
        \mathbb{E}_{\xi}\left[
            (f^2_w)^2
        \right] \\
        \bar{V} = (1+\gamma) \mathbb{E}_{\xi}\left[
            \mathcal{Z}_{w_\star}
            \partial_\varrho f^1_w
        \right]
        + (1-\gamma)
        \mathbb{E}_{\xi}\left[
            \partial_\varrho f^2_w
        \right] \\
        \bar{P} = \gamma \mathbb{E}_{\xi}\left[
            \mathcal{Z}_{w_\star}
            \abs{f^1_w}^{\defnorm^\star}
        \right]
        + (1-\gamma)
        \mathbb{E}_{\xi}\left[
            \abs{f^2_w}^{\defnorm^\star}
        \right]
    \end{cases}
    &\text{ for } & \gamma \leq 1 \,, \label{eq:app:latent-space-eq-prior-gamma-small}\\
    &\begin{cases}
        \bar{m} = \frac{1}{\sqrt{\gamma}} \mathbb{E}_{\xi}\left[
            \partial_\varrho \mathcal{Z}_{w_\star}
            f_w
        \right]\\
        \bar{\tau} = 2 \mathbb{E}_{\xi}\left[
            \mathcal{Z}_{w_\star}
            f_w
        \right] \\
        \bar{V} = 2 \mathbb{E}_{\xi}\left[
            \mathcal{Z}_{w_\star}
            \partial_\varrho f_w
        \right] \\
        \bar{P} = \mathbb{E}_{\xi}\left[
            \mathcal{Z}_{w_\star}
            \abs{f_w}^{\defnorm^\star}
        \right]
    \end{cases}
    & \text{ for } &  \gamma > 1 \,, \label{eq:app:latent-space-eq-prior-gamma-big}
\end{align}

depends on the regularization norm and the prior over the ground truth weights through
$\mathcal{Z}_{w_\star} \equiv \mathcal{Z}_{w_\star} (\sfrac{\bar{m}\xi}{\sqrt{(1+\gamma)\bar{\tau}}}, \sfrac{\bar{m}^2}{(1+\gamma) \bar{\tau}})$,
$f_w^1 \equiv f_w (\sqrt{\bar{\tau}} \sqrt{1 + \frac{1}{\gamma}} \xi, \bar{V}(1 + \frac{1}{\gamma}), \sfrac{\bar{P}}{2})$ and
$f_w^2 \equiv f_w (\sqrt{\bar{\tau}}\xi, \bar{V}, \sfrac{\bar{P}}{2})$
for $\gamma\leq 1$ and
$\mathcal{Z}_{w_\star} \equiv \mathcal{Z}_{w_\star} (\sfrac{\bar{m}\xi}{\sqrt{2 \bar{\tau}}}, \sfrac{\bar{m}^2}{2 \bar{\tau}})$ and
$f_w \equiv f_w (\sqrt{2 \bar{\tau}}\xi, 2\bar{V}, \sfrac{\bar{P}}{2})$ for $\gamma > 1$ where
\begin{equation}
    f_w(\varrho, \Lambda, \pi) = \argmin_{z} \qty[ \revision{\pi\, |z|^{\defnorm^\star}} + \lambda z^2 + \frac{\Lambda}{2} z^2 - \varrho z]\,.
\end{equation}

\subsubsection{Specialization for $s=2$}

To obtain the specific form of the saddle-point equations presented in~\Cref{thm:self-consistent-equations-latent-space}, one applies the closed-form proximal operator for $s=2$,
\begin{equation}
    f_w(\varrho, \Lambda, \pi) = \frac{\varrho}{\lambda + \pi + \Lambda}\,.
\end{equation}
This renders~\Cref{eq:latent-space-eq-prior-gamma-small,eq:latent-space-eq-prior-gamma-big} amenable to explicit integration: they become Gaussian integrals over the variable $\xi$. Performing these Gaussian integrals yields the form shown in the main text.

\subsection{Limiting Values for the Error Metrics}\label{sec:app:limiting-error-metrics}

We start by proving the following lemma that will be useful in the following.

{
\begin{lemma}[Consistent perturbation minimization]\label{lemm:adv-decreasing-fun}
For any decreasing function $g\colon\RR\to\RR$, any $y\in\qty{\pm 1}$, and any vectors $\x,\w \in \RR^d$,
\begin{equation}\label{eq:app:lemm:minimization}
    \max_{
        \attack : \norm{\attack}_{\attnorm} \leq \varepsilon, \langle \wstar, \attack \rangle = 0
    }
    g\qty(y\expval{ \w, \x + \attack})
    =
    \inf_{\kappa \in\RR}g\qty(y\expval{ \w, \x} - \varepsilon \norm{\w - \kappa \wstar}_\qstar)
\end{equation}
\end{lemma}
}

\begin{proof}[Proof of \Cref{lemm:adv-decreasing-fun}]
Since $g : \RR \to \RR$ in~\cref{eq:app:lemm:minimization} is decreasing (but not necessarily continuous), it suffices to minimize its argument and then evaluate $g$ at the minimizer. Splitting the argument,
\begin{equation}
    \min_{
        \attack : \norm{\attack}_{\attnorm} \leq \varepsilon,\, \langle \wstar, \attack \rangle = 0
    }
    y\,\expval{ \w, \x + \attack }
    = y\,\expval{ \w, \x }
    + \min_{\attack : \norm{\attack}_{\attnorm} \leq \varepsilon,\, \langle \wstar, \attack \rangle = 0}
    y\,\expval{ \w, \attack }\,,
\end{equation}
the first term does not depend on $\attack$, so we focus on the second.
Since $y \in \qty{+1,-1}$, the change of variables $\attack \to y\attack$ leaves the constraints invariant (the $\attnorm$-norm is even and the orthogonality constraint is linear) and removes the $y$ factor from the objective. We can therefore write
\begin{equation}
    \min_{\attack : \norm{\attack}_{\attnorm} \leq \varepsilon,\, \langle \wstar, \attack \rangle = 0} \langle \w, \attack \rangle
    = \min_{\attack : \norm{\attack}_{\attnorm} \leq \varepsilon} \sup_{\kappa\in\RR} \langle \w, \attack \rangle + \kappa \langle \wstar, \attack \rangle
    = \min_{\attack : \norm{\attack}_{\attnorm} \leq \varepsilon} \sup_{\kappa\in\RR} \langle \w + \kappa \wstar, \attack \rangle\,,
\end{equation}
where we introduced a Lagrange multiplier $\kappa$ for the linear constraint. Strong duality holds for this min--sup, so we may interchange the order:
\begin{equation}
    \sup_{\kappa\in\RR} \min_{\attack : \norm{\attack}_{\attnorm} \leq \varepsilon} \langle \w + \kappa \wstar, \attack \rangle
    = \sup_{\kappa\in\RR} \bigl(- \varepsilon \norm{\w + \kappa \wstar}_{\qstar}\bigr)
    = - \varepsilon \inf_{\kappa\in\RR} \norm{\w + \kappa \wstar}_{\qstar}\,,
\end{equation}
using the duality $\sfrac{1}{\attnorm} + \sfrac{1}{\qstar} = 1$. \revision{Hence the optimal attack contribution is}
\begin{equation}\label{eq:app:minimal-norm}
    \revision{- \varepsilon \inf_{\kappa\in\RR} \norm{\w + \kappa \wstar}_{\qstar}}\,,
\end{equation}
which is the term that appears inside $g$ on the right-hand side of~\cref{eq:app:lemm:minimization}.
\end{proof}

\subsubsection{Proper Error Metrics}

We now compute the limiting values of the proper error metrics.
\Cref{lemm:adv-decreasing-fun} is the basis of the proofs of~\Cref{thm:metrics-latent-space,thm:metrics-well-spec}: it tells us that, given the joint limiting distribution of $\wstar$ and $\w$, one can evaluate the limiting form of the attack term as a function of $\kappa$ and then take the extremization over $\kappa$.

\begin{proof}[Proof of~\Cref{thm:metrics-well-spec}]
We first recall that the limiting distribution of the trained estimator is
\begin{equation}\label{eq:limining-w-dist}
    \what \sim \mathcal{P}_{\bar{V} \tilde{r}(\cdot, \bar{P})} \qty(\bar{m} \wstar + \sqrt{\bar{\tau} - \bar{m}^2} \boldsymbol{\xi})\,,
\end{equation}
where $\bar{m}$, $\bar{\tau}$, $\bar{V}$ and $\bar{P}$ are the solution of the saddle-point equations of~\citet{vilucchio2024geometry}. In the case $s=2$ the proximal can be evaluated in closed form, yielding $\what_i = m (\wstar)_i + \sqrt{\tau - m^2}\,\xi_i$ with $\xi \sim \mathcal{N}(\boldsymbol{0}, \idmat_d)$, where $m$ and $\tau$ are the (unbarred) order parameters determined by the saddle point.

We now evaluate the limit of the term in~\cref{eq:app:minimal-norm}, which is the $\ell^{q^{\star}}$-norm of a Gaussian vector with a specific covariance.
For any $p\in[1,\infty]$ the $\ell^{p}$-norm of a Gaussian vector concentrates around its expectation: this is an application of \citep[Theorem~5.5]{vershynin2018high} to the Lipschitz function $\norm{M\,\cdot}_{p}$, where $M$ is the square root of the covariance of the Gaussian vector.
The explicit limit for our case can be found as \citep[Corollary~1]{biau2015high}.
\revision{Under the scaling $\tilde{\varepsilon} \coloneqq \varepsilon\,d^{1/q^{\star}}$ of the adversarial budget, we obtain}
\begin{equation}
    \revision{\varepsilon \inf_{\kappa} \norm{\tilde{\w} + \kappa \wstar}_{q^\star}}
    \xrightarrow[d\to\infty]{}
    \tilde{\varepsilon} \inf_{\kappa}
    \frac{\sqrt{2}}{\pi^{(2 q^\star)^{-1}}}
    \sqrt{\qty(m + \kappa)^2 + \tau - m^2}
    \sqrt[q^\star]{\Gamma\qty(\frac{q^\star + 1}{2})}
\end{equation}
the previous equation is always minimized for $\kappa = -m$ and thus it leads to
\begin{equation}
    \tilde{\varepsilon}
    \sqrt{2}
    \sqrt{\tau - m^2}
    \sqrt[q^\star]{\frac{\Gamma\qty(\sfrac{(q^\star + 1)}{2})}{\sqrt{\pi}}} \,.
\end{equation}
The existence of such a closed form is a peculiarity of the case $s=2$. In the general case the limiting value still exists because the proximal of a convex function is itself Lipschitz; applying~\cref{eq:limining-w-dist} then yields
\begin{equation}
    \mathcal{A} = \inf_{\kappa \in \RR} \mathbb{E}_{\xi}\qty[ \int \dd{w_\star} \abs{ f_{w} - \kappa w_\star }^{\qstar} ]^{1/\qstar} \,.
\end{equation}
where $f_w = \mathcal{P}_{\bar{V} \tilde{r}(\cdot, \bar{P})} \qty(\bar{m} w_\star + \sqrt{\bar{\tau} - \bar{m}^2} \xi)$ is the limiting distribution of the entries of $\what$.

Once the attack-surface term concentrates to its limit, the limiting distribution of the metrics~\cref{eq:boundary-proper,eq:def-misclassif-proper} can be obtained by the \textit{local fields} method \citep{clarte_double_descent}: in the well-specified model the relevant pair of random variables is
\begin{equation}
    \begin{pmatrix}
        \revision{\langle\wstar , \x \rangle} \\
        \revision{\langle\what , \x \rangle}
    \end{pmatrix}
    \sim
    \mathcal{N}
    \qty(
    \begin{pmatrix} 0 \\ 0 \end{pmatrix},
    \begin{pmatrix}
        1 & m \\
        m & \tau
    \end{pmatrix}
    )\,,
\end{equation}
where $m$ and $\tau$ are the order parameters obtained from the saddle-point equations.
{
For completeness, let $\mathsf P(\dd y\mid z_1)$ denote the conditional
label law on the label space $\mathcal Y$, and set
$a=\tilde\varepsilon\mathcal A\geq0$. If $\attack_0$ minimizes the raw model
field, the change of variables in the proof of~\Cref{lemm:adv-decreasing-fun}
shows that a sample with label $y$ receives $y\attack_0$. Consequently,
$z_2^{\rm adv}=z_2-ya$ and its signed margin is
$yz_2^{\rm adv}=yz_2-a$. In particular, the attack subtracts $a$ from the
\emph{signed margin}; the expression $y(z_2-a)$ would instead apply the same
raw-field shift to both labels and is not equivalent. Applying the lemma to
the indicator loss therefore gives the generic-channel formulas
\begin{align}
\adverrproper
&=\int\dd p(z_1,z_2)\int_{\mathcal Y}\mathsf P(\dd y\mid z_1)
\mathbbm{1}\{yz_2<a\},\label{eq:app:generic-rob-proper}\\
\bounderrproper
&=\int\dd p(z_1,z_2)\int_{\mathcal Y}\mathsf P(\dd y\mid z_1)
\mathbbm{1}\{0<yz_2<a\}.\label{eq:app:generic-bnd-proper}
\end{align}
Since $a\geq0$, these indicators also obey, away from the zero-probability
Gaussian boundaries,
\begin{equation}\label{eq:app:rob-clean-bnd-identity}
\mathbbm{1}\{yz_2<a\}
=\mathbbm{1}\{yz_2<0\}+\mathbbm{1}\{0<yz_2<a\}.
\end{equation}
Thus $\adverrproper$ is the clean error plus $\bounderrproper$, providing a
direct consistency check on the formulas.
For binary labels the inner integral is the corresponding sum over
$y\in\{\pm1\}$. For a noiseless link, the conditional law is the point mass
at $y=\sign(z_1)$, and these expressions
reduce to~\cref{eq:well-specified-rob-proper,eq:well-specified-bnd-proper}.
The same signed-margin calculation gives the standard robust error after
replacing $\mathcal A$ by $\mathcal B$, and the latent-space formulas after
replacing $\mathcal A$ by $\mathcal C$.
}
\end{proof}

\begin{proof}[Proof of~\Cref{thm:metrics-latent-space}]
The argument mirrors the proof of \Cref{thm:metrics-well-spec}; only the
limiting form of the attack-surface constant changes, and that change is
governed by the block structure of the feature matrix $\featmat$
(\Cref{asm:data-distr-latent-space}).

\textit{Step 1: Apply~\Cref{lemm:adv-decreasing-fun} in the latent
attack form.}
By~\Cref{prop:exist:latent}, a consistent latent-space perturbation
$\attack\in\mathbb{R}^{d}$ with $\langle\wstar,\attack\rangle=0$ acts on the
model field through $\langle\thetahat,\featmat\attack\rangle$. Applying
\Cref{lemm:adv-decreasing-fun} with $\w\leftarrow\featmat^{\top}\thetahat$
and $\x\leftarrow\z$, for any non-increasing $g$,
\begin{equation}\label{eq:app:latent-lemma}
    \revision{\max_{\attack\in H_{\attnorm}(\varepsilon)} g\!\qty(y\langle\thetahat,\featmat(\z+\attack)+\uu\rangle)
    =\inf_{\kappa\in\mathbb{R}} g\!\qty(y\langle\thetahat,\featmat\z+\uu\rangle
    -\varepsilon\norm{\featmat^{\top}\thetahat-\kappa\wstar}_{\qstar})}.
\end{equation}
Under~\Cref{asm:scaling-eps-well-spec}, the attack contribution
concentrates to $\tilde{\varepsilon}\,\mathcal{C}$, with
\begin{equation*}
    \mathcal{C} \;=\; \lim_{d,p,n\to\infty}\;
    \revision{\tfrac{1}{d^{1/\qstar}}}
    \inf_{\kappa\in\mathbb{R}}\norm{\featmat^{\top}\thetahat-\kappa\wstar}_{\qstar}.
\end{equation*}

\textit{Step 2: Block decomposition of $\featmat^{\top}\thetahat$.}
\Cref{asm:data-distr-latent-space} prescribes
\begin{equation*}
\revision{\featmat=\begin{cases}
\left[\idmat_{d};\;\boldsymbol{0}_{(p-d)\times d}\right] & p\geq d,\\
\sqrt{\tfrac{d}{p}}\left[\idmat_{p},\;\boldsymbol{0}_{p\times(d-p)}\right] & p<d.
\end{cases}}
\end{equation*}
Hence, writing $\thetahat=(\thetahat^{\rm s},\thetahat^{\rm k})$ for the
block decomposition induced by $\featmat$:
\begin{itemize}
    \item \revision{For $\gamma\leq 1$ ($p\geq d$): $\thetahat^{\rm s}=\thetahat_{1:d}\in\mathbb{R}^{d}$ couples to $\wstar$ through $\featmat^{\top}\thetahat=\thetahat^{\rm s}$; the kernel block $\thetahat_{d+1:p}$ does not contribute to the latent attack surface.}
    \item \revision{For $\gamma>1$ ($p<d$): $\featmat^{\top}\thetahat=\sqrt{\gamma}(\thetahat^{\top},\boldsymbol{0}_{d-p}^{\top})^{\top}$, so the first $p$ coordinates carry the signal and the remaining $d-p$ coordinates are zero.}
\end{itemize}
The rewriting of the saddle-point equations in~\Cref{sec:app:proof-high-d} (\Cref{eq:latent-space-eq-prior-gamma-small,eq:latent-space-eq-prior-gamma-big}) implements precisely this decomposition: it yields the limiting laws
$f_{w}^{1}$ (signal block, $\gamma\leq 1$), $f_{w}^{2}$ (kernel block,
$\gamma\leq 1$) and $f_{w}^{3}$ (single block, $\gamma>1$) for the
empirical distribution of the entries of $\thetahat$, expressed through
the proximal of the regularizer.

\textit{Step 3: Limit of the dual norm.}
The $\ell^{\qstar}$-norm of a vector with i.i.d.\ asymptotically Gaussian
entries concentrates by the Lipschitz-Gaussian inequality
(\citet[Theorem~5.5]{vershynin2018high}; cf.\ the well-specified argument).
Splitting the sum that defines
$\norm{\featmat^{\top}\thetahat-\kappa\wstar}_{\qstar}^{\qstar}$ across the
blocks identified in Step~2 and using the empirical-distribution limits
from \citet[Lemma~5]{loureiro2021learning} yields
\begin{equation}\label{eq:app:C-value}
    \revision{\mathcal{C} \;=\; \inf_{\kappa\in\mathbb{R}}
    \begin{cases}
    \Big(
        \mathbb{E}_{\xi}\!\!\int\!\! \dd w_\star\, h^{1}(w_\star)\,\abs{f_{w}^{1}-\kappa w_{\star}}^{\qstar}
    \Big)^{1/\qstar}
    & \gamma\leq 1,\\[6pt]
    \Big(
        \frac{1}{\gamma}\mathbb{E}_{\xi}\!\!\int\!\! \dd w_\star\, h^{3}(w_\star)\,\abs{\sqrt{\gamma}f_{w}^{3}-\kappa w_{\star}}^{\qstar}
        +\left(1-\frac{1}{\gamma}\right)\mathbb{E}_{\xi}\!\!\int\!\! \dd w_\star\, h^{3}(w_\star)\,\abs{\kappa w_{\star}}^{\qstar}
    \Big)^{1/\qstar}
    & \gamma>1,
    \end{cases}}
\end{equation}
where the densities $h^{i}(w_\star)$ encode the Gaussian "channel" prior on $w_{\star}$ at the relevant saddle point and the $f_{w}^{i}$ are the limiting proximals of Step~2. Explicitly, with
\begin{equation*}
    h(w_{\star};\varrho,\Lambda)\;=\;e^{-\sfrac{w_{\star}^{2}}{2}}\,e^{-\sfrac{\Lambda w_{\star}^{2}}{2}\,+\,\varrho\, w_{\star}},
\end{equation*}
\revision{one has $h^{1}\!=\!h(w_{\star};\sfrac{\bar{m}\xi}{\sqrt{(1+\gamma)\bar{\tau}}},\sfrac{\bar{m}^{2}}{(1+\gamma)\bar{\tau}})$ and $h^{3}\!=\!h(w_{\star};\sfrac{\bar{m}\xi}{\sqrt{2\bar{\tau}}},\sfrac{\bar{m}^{2}}{2\bar{\tau}})$. In the $\gamma>1$ branch, the weights $1/\gamma=p/d$ and $1-1/\gamma$ are the fractions of nonzero and zero coordinates of $\featmat^\top\thetahat$, respectively.}

\textit{Step 4: Local-fields conclusion.}
\revision{Substituting $\mathcal{C}$ into~\eqref{eq:app:latent-lemma} and applying
the generic-channel formulas~\cref{eq:app:generic-rob-proper,eq:app:generic-bnd-proper}
with $\mathcal A$ replaced by $\mathcal C$}
with \revision{$z_{1}=\langle\wstar,\z\rangle$ and
$z_{2}=\langle\thetahat,\x\rangle$}
(jointly Gaussian with the covariance stated in~\Cref{thm:metrics-latent-space},
by~\citet[Lemma~5]{loureiro2021learning}) yields
\revision{the generic noisy-channel result. For the noiseless link $y=\sign(z_1)$, it reduces to
\eqref{eq:latent-space-rob-proper} and \eqref{eq:latent-space-bnd-proper}.}
\end{proof}

\begin{corollary}[Euclidean simplification of $\mathcal{C}$]
\label{cor:C-q-equals-2}
Under the setting of~\Cref{thm:metrics-latent-space}, when the attack norm is Euclidean ($\attnorm=2$, hence $\qstar=2$),
the attack-surface constant in~\Cref{eq:app:C-value} simplifies to
\begin{equation}\label{eq:app:C-l2}
    \mathcal{C} \;=\; \sqrt{\qlatent - m^{2}},
\end{equation}
where $\qlatent \coloneqq \sfrac{\norm{\featmat^{\top}\thetahat}_{2}^{2}}{d}$ is the latent-space self-overlap and $m \coloneqq \sfrac{\langle\featmat^{\top}\thetahat,\wstar\rangle}{d}$ is the target alignment, both as in~\Cref{thm:self-consistent-equations-latent-space}.
\end{corollary}

\begin{proof}
For $\qstar=2$, the dual norm appearing in~\eqref{eq:app:latent-lemma} is Euclidean, and the infimum admits a one-dimensional Hilbert-space projection in closed form. For any $v\in\RR^{d}$ and any nonzero $\wstar\in\RR^{d}$,
\begin{equation}\label{eq:app:hilbert-projection}
    \inf_{\kappa\in\RR}\norm{v-\kappa\wstar}_{2}^{2}
    \;=\;\norm{v}_{2}^{2}-\frac{\langle v,\wstar\rangle^{2}}{\norm{\wstar}_{2}^{2}},
    \qquad
    \text{attained at }\kappa^{\star}=\frac{\langle v,\wstar\rangle}{\norm{\wstar}_{2}^{2}},
\end{equation}
which is Pythagoras applied to the orthogonal decomposition
$v=\frac{\langle v,\wstar\rangle}{\norm{\wstar}^{2}}\wstar + P_{\perp}v$
with $P_{\perp}=\idmat_{d}-\sfrac{\wstar\wstar^{\top}}{\norm{\wstar}^{2}}$.

Specializing to $v=\featmat^{\top}\thetahat$, and using $\norm{\wstar}_{2}^{2}=d$ from $\wstar\in\mathbb{S}^{d-1}(\sqrt{d})$,
\begin{equation*}
    \inf_{\kappa\in\RR}\norm{\featmat^{\top}\thetahat-\kappa\wstar}_{2}^{2}
    \;=\;\norm{\featmat^{\top}\thetahat}_{2}^{2}-\frac{\langle \featmat^{\top}\thetahat,\wstar\rangle^{2}}{d}
    \;=\;d\,\qlatent-\frac{(d\,m)^{2}}{d}
    \;=\;d\,(\qlatent-m^{2}),
\end{equation*}
attained at $\kappa^{\star}=m$. Substituting into the definition of $\mathcal{C}$ in Step~3 of the proof of~\Cref{thm:metrics-latent-space} — for $\qstar=2$ the dimensional prefactor exactly cancels the $\sqrt{d}$ inside the square root — gives~\eqref{eq:app:C-l2}.

\revision{This Euclidean projection does not require the case split in~\eqref{eq:app:C-value}: by the definitions of $\qlatent$, $m$, and $\|\wstar\|_2^2/d=1$, expanding the quadratic in either branch reduces it to}
\begin{equation*}
    \mathcal{C}^{2}\;=\;\inf_{\kappa\in\RR}\left[\qlatent-2\kappa m+\kappa^{2}\right]\;=\;\qlatent-m^{2},
\end{equation*}
confirming~\eqref{eq:app:C-l2} directly from~\eqref{eq:app:C-value}.
\end{proof}

\subsection{Asymptotic Behavior Under Extreme Overparameterization}\label{sec:app:extreme-overparam}

This subsection is devoted to the proof of Proposition~\ref{prop:extreme-overparam}.
We analyze the behavior of the consistent errors in the limit $\psi \to \infty$, which corresponds to $\gamma = d/p \to 0^+$ at fixed sample complexity $\alpha$ and adversarial budget $\varepsilon$.

\begin{proof}[Proof of Proposition~\ref{prop:extreme-overparam}]
{
Put $t=\sqrt\gamma$ and define, for later use, the standard Gaussian density
and distribution function by
\begin{equation*}
\phi(u)=\frac{e^{-u^2/2}}{\sqrt{2\pi}},
\qquad
\Phi(v)=\int_{-\infty}^{v}\phi(u)\,\dd u.
\end{equation*}

\textit{Step 1: Scaling of the saddle-point solution.}
We first justify the expansion of the order parameters from
\cref{eq:latent-space-eq-channel,eq:latent-space-eq-prior-gamma-small}.
For the logistic loss $\ell(y,z)=\log(1+e^{-yz})$, the scalar proximal has
the representation~\citep[Proposition~2]{Briceno-Arias2019}
\begin{equation}\label{eq:app:logistic-prox-lambert}
\mathcal P_{V\ell(y,\cdot)}(\omega)
=\omega+yW_{e^{-y\omega}}\!\left(Ve^{-y\omega}\right),
\end{equation}
where the generalized Lambert function $W_r$ is defined by
$u(e^u+r)=v$ if and only if $u=W_r(v)$. Together with the proximal shift
identity~\eqref{eq:app:proximal-moreau-shift}, this also covers the shifted
loss in the robust-training residual $f_\lossfun(y,\omega,V,P)$.
Equivalently, the proximal optimality condition gives
\begin{equation*}
\abs{f_\lossfun(y,\omega,V,P)}<1,
\qquad
\partial_\omega f_\lossfun(y,\omega,V,P)<0,
\end{equation*}
and both quantities are smooth in $(\omega,V,\sqrt P)$ for $V>0$.
We therefore expand the fixed-point equations in $t$, following the
asymptotic fixed-point methodology used for high-dimensional logistic
ERM~\citep[Supplementary Sec.~V]{aubin_generalization_2020}.

For a noiseless link, if $\eta=m/\sqrt\tau$, the channel appearing in
\cref{eq:latent-space-eq-channel} is explicitly
\begin{equation}\label{eq:app:extreme-noiseless-channel}
\mathcal Z_0(y,\eta\xi,1-\eta^2)
=\Phi\!\left(\frac{y\eta\xi}{\sqrt{1-\eta^2}}\right).
\end{equation}
In particular, at $\eta=0$,
$\mathcal Z_0(y,0,1)=1/2$ and
$\partial_\omega\mathcal Z_0(y,0,1)=y\phi(0)$.
The explicit powers of $t$ in the channel equations and the boundedness
above first give
\begin{equation}\label{eq:app:extreme-preliminary-scaling}
\bar m=O(t),\qquad \bar\tau=O(t^2),\qquad
\bar V=O(t^2),\qquad \bar P=O(\sqrt P).
\end{equation}
Let $D_\gamma=2\lambda+\bar P+\bar V(1+\gamma^{-1})$ be the denominator
in~\cref{eq:latent-space-eq-prior-gamma-small}. Here regularity means that
$D_\gamma$ stays bounded away from zero, the bounded hyperparameters are
$C^2$ in $t$ with $\lambda=\lambda_0+O(t)$, and the limiting saddle system
has a nonsingular Jacobian. The prior equation for
$P$ and~\eqref{eq:app:extreme-preliminary-scaling} then imply
$P=O(t^2)$, so $\bar P=O(t)$. The remaining prior equations consequently
give $m=O(t)$, $\tau=O(1)$, and $V=O(1)$.
The leading coefficients can also be read directly from the equations.
Define
\begin{align*}
f_0(y,\xi)&=f_\lossfun(y,\sqrt{\tau_0}\xi,V_0,0),\\
M_0&=\mathbb E_\xi\!\left[\sum_{y=\pm1}y\phi(0)f_0(y,\xi)\right],\\
T_0&=\mathbb E_\xi\!\left[\frac12\sum_{y=\pm1}f_0(y,\xi)^2\right],\\
S_0&=-\mathbb E_\xi\!\left[\frac12\sum_{y=\pm1}
\partial_\omega f_\lossfun(y,\sqrt{\tau_0}\xi,V_0,0)\right].
\end{align*}
Since $y f_0(y,\xi)>0$ and $\partial_\omega f_\lossfun<0$ for logistic
loss, $M_0,T_0,S_0>0$. Expanding
\eqref{eq:app:extreme-noiseless-channel} and the smooth proximal residual
in $t$ yields
\begin{equation}\label{eq:app:extreme-barred-scaling}
\bar m=\alpha M_0t+O(t^2),\quad
\bar\tau=\alpha T_0t^2+O(t^3),\quad
\bar V=\alpha S_0t^2+O(t^3),\quad
\bar P=O(t).
\end{equation}
Thus the limiting saddle system is
\begin{equation}\label{eq:app:extreme-limiting-saddle}
\begin{gathered}
D_0=2\lambda_0+\alpha S_0,\qquad V_0=D_0^{-1},\\
m_0=\frac{\alpha M_0}{D_0},\qquad
\tau_0=\frac{\alpha^2M_0^2+\alpha T_0}{D_0^2},\qquad
P_0=\frac{\alpha^2M_0^2+2\alpha T_0}{D_0^2}.
\end{gathered}
\end{equation}
These relations are self-consistent because $M_0,T_0,S_0$ depend on
$(\tau_0,V_0)$. The robust-training budget $r=O(1)$ enters through
$\bar P=O(t)$ and hence only at the next order. The implicit-function
theorem applied to~\eqref{eq:app:extreme-limiting-saddle} gives
\begin{equation}\label{eq:app:extreme-unbarred-scaling}
m=m_0t+O(t^2),\qquad \tau=\tau_0+O(t),\qquad
P=P_0t^2+O(t^3),\qquad V=V_0+O(t).
\end{equation}
In particular, $m_0>0$ on the positively aligned branch and $\tau_0>0$.

\textit{Step 2: Expansion of the attack threshold.}
For $q=2$, Corollary~\ref{cor:C-q-equals-2} and the definition
$\tau=\qlatent+P$ give the exact identity
\begin{equation}\label{eq:app:extreme-C-exact}
\mathcal C^2=\tau-P-m^2.
\end{equation}
Define the normalized alignment and attack threshold
\begin{equation*}
\eta_\gamma=\frac{m}{\sqrt\tau},\qquad
b_\gamma=\frac{\tilde\varepsilon\mathcal C}{\sqrt\tau},\qquad
k_\gamma=\frac{\eta_\gamma}{\sqrt{1-\eta_\gamma^2}}.
\end{equation*}
Equations~\eqref{eq:app:extreme-unbarred-scaling}--\eqref{eq:app:extreme-C-exact}
imply
\begin{equation}\label{eq:app:extreme-normalized-scaling}
k_\gamma=\frac{m_0}{\sqrt{\tau_0}}\sqrt\gamma+O(\gamma),
\qquad b_\gamma=\tilde\varepsilon+O(\gamma).
\end{equation}

\textit{Step 3: Expansion of the error metrics.}
Let $U=z_2/\sqrt\tau$. Since $(z_1,U)$ is standard bivariate Gaussian with
correlation $\eta_\gamma$, conditioning on $U=u$ gives
$\mathbb P(z_1>0\mid U=u)=\Phi(k_\gamma u)$. The noiseless formulas
in~\Cref{thm:metrics-latent-space} and the central symmetry of
$(z_1,z_2)$ therefore give
\begin{align}
\adverrproper
&=2\,\mathbb P(z_1>0,z_2<\tilde\varepsilon\mathcal C)
=2\int_{-\infty}^{b_\gamma}\phi(u)\Phi(k_\gamma u)\,\dd u,
\label{eq:app:extreme-rob-integral}\\
\bounderrproper
&=2\,\mathbb P(z_1>0,0<z_2<\tilde\varepsilon\mathcal C)
=2\int_{0}^{b_\gamma}\phi(u)\Phi(k_\gamma u)\,\dd u.
\label{eq:app:extreme-bnd-integral}
\end{align}
The Taylor expansion
$\Phi(ku)=\tfrac12+\phi(0)ku+O(k^3|u|^3)$ is integrable against
$\phi(u)\dd u$. Using
\begin{equation*}
\int_{-\infty}^{b}u\phi(u)\,\dd u=-\phi(b),
\qquad
\int_0^b u\phi(u)\,\dd u=\phi(0)-\phi(b),
\end{equation*}
together with~\eqref{eq:app:extreme-normalized-scaling} gives
\begin{align}
\adverrproper
&=\Phi(\tilde\varepsilon)
-\frac{m_0}{\pi\sqrt{\tau_0}}e^{-\tilde\varepsilon^2/2}\sqrt\gamma
+O(\gamma),\label{eq:app:extreme-rob-expansion}\\
\bounderrproper
&=\Phi(\tilde\varepsilon)-\frac12
+\frac{m_0}{\pi\sqrt{\tau_0}}
\left(1-e^{-\tilde\varepsilon^2/2}\right)\sqrt\gamma
+O(\gamma).\label{eq:app:extreme-bnd-expansion}
\end{align}
Regularity of the saddle branch makes these remainders differentiable.
Finally, $\gamma=(\alpha\psi)^{-1}$ implies
$\partial_\psi=-\alpha\gamma^2\partial_\gamma$; differentiating
\cref{eq:app:extreme-rob-expansion,eq:app:extreme-bnd-expansion} yields
\begin{align*}
\partial_\psi\adverrproper
&=\frac{\alpha m_0}{2\pi\sqrt{\tau_0}}
e^{-\tilde\varepsilon^2/2}\gamma^{3/2}+O(\gamma^2),\\
\partial_\psi\bounderrproper
&=-\frac{\alpha m_0}{2\pi\sqrt{\tau_0}}
\left(1-e^{-\tilde\varepsilon^2/2}\right)\gamma^{3/2}
+O(\gamma^2),
\end{align*}
which are~\cref{eq:extreme-rob-derivative,eq:extreme-bnd-derivative}.
The first leading coefficient is positive. The second is non-positive,
and it is strictly negative for $\tilde\varepsilon>0$; at
$\tilde\varepsilon=0$, the boundary error is identically zero.}
\end{proof}

\section{Replica derivation of the self-consistent equations}\label{sec:app:replica-computation}

In this appendix we give an alternative derivation of the asymptotic equations of~\Cref{thm:self-consistent-equations-latent-space} using the (heuristic) replica method from statistical physics \citep{mezard1987spin}. The replica computation reproduces \emph{exactly} the self-consistent equations~\cref{eq:latent-space-eq-channel,eq:latent-space-eq-prior-gamma-small,eq:latent-space-eq-prior-gamma-big} obtained from the rigorous CGMT-based derivation of~\Cref{sec:app:proof-high-d}, providing an independent cross-check. We sketch the main steps and refer to \cite{loureiro2021learning, vilucchio2025asymptotics} for the detailed manipulations, which are standard in the high-dimensional asymptotics literature.

\subsection{Gibbs measure and free energy}

To the adversarial training problem~\cref{eq:risk-definition,eq:risk-minimization} we associate the Gibbs measure at inverse temperature $\beta>0$,
\begin{equation}\label{eq:replica:gibbs}
    \mu_\beta(\dd\w) = \frac{1}{\mathcal{Z}_\beta}\exp\!\left\{-\beta\sum_{\mu=1}^{n}\lossfun\!\left(y_\mu\frac{\w^{\top}\x_\mu}{\sqrt{p}} - \frac{\advtrainingcost}{\sqrt{p}}\norm{\w}_{\qstar}\right) - \frac{\beta\lambda}{2}\norm{\w}_{2}^{2}\right\}\dd\w\,,
\end{equation}
whose partition function $\mathcal{Z}_\beta$ concentrates on the minimizers of the adversarially trained risk as $\beta\to\infty$. The quantity of interest is the dataset-averaged free-energy density
\begin{equation}\label{eq:replica:free-energy}
    f := -\lim_{\beta\to\infty}\lim_{d\to\infty}\frac{1}{\beta\,d}\,\mathbb{E}_{\mathcal{D}}\log\mathcal{Z}_\beta\,,
\end{equation}
which we compute via the replica trick $\mathbb{E}_{\mathcal D}\log\mathcal{Z}_\beta = \lim_{r\to 0^{+}}\partial_{r}\mathbb{E}_{\mathcal D}\mathcal{Z}_\beta^{r}$.

\subsection{Replicated partition function and order parameters}

After averaging over the data, the $r$-replicated partition function depends on the linear combinations $\nu_\mu := \langle\wstar,\z_\mu\rangle/\sqrt{d}$ and $\lambda_\mu^{a} := \langle\w^{a},\x_\mu\rangle/\sqrt{p}$ (for $a=1,\dots,r$), which are jointly Gaussian with the covariances
\begin{equation}\label{eq:replica:overlaps}
    \mathbb{E}\,\nu_\mu^{2} = 1\,,\quad
    \mathbb{E}\,\lambda^{a}_\mu\nu_\mu = \frac{\sqrt{\gamma}}{d}\,\wstar^{\top}\featmat^{\top}\w^{a}\,,\quad
    \mathbb{E}\,\lambda^{a}_\mu\lambda^{b}_\mu = \frac{1}{p}\,(\w^{a})^{\top}(\featmat\featmat^{\top}+\idmat_{p})\,\w^{b}\,.
\end{equation}
Introducing the order parameters $m^{a}$, $Q^{ab}$, $P^{a}$ for these covariances together with the associated Lagrange multipliers $\hat m^{a}$, $\hat Q^{ab}$, $\hat P^{a}$ via Fourier representations of the constraining $\delta$-functions, the replicated partition function takes the saddle-point form
\begin{equation}\label{eq:replica:phi-decomposition}
    \mathbb{E}_{\mathcal{D}}\mathcal{Z}_\beta^{r} \;\propto\; \int\dd\Theta\,\dd\hat\Theta\;\exp\!\bigl\{p\,\Phi^{(r)}(\Theta,\hat\Theta)\bigr\}\,,\qquad
    \Phi^{(r)} = \Psi_{t} + \alpha\gamma\,\Psi_{y}^{(r)} + \Psi_{w}^{(r)}\,,
\end{equation}
where $\Psi_{t}$ is the Legendre-transform (trace) term, and $\Psi_{y}^{(r)},\Psi_{w}^{(r)}$ are the channel and prior contributions, whose explicit forms follow standard manipulations \citep{mezard1987spin, loureiro2021learning} and are omitted here.

\subsection{Replica-symmetric ansatz and zero-temperature limit}

We adopt the replica-symmetric (RS) ansatz $m^{a}=m$, $Q^{aa}=V+q$, $Q^{ab}=q$ for $a<b$, $P^{a}=P$, together with matching scalings for the hatted variables, and take the zero-temperature limit through the standard substitutions $V\to V/\beta$, $\hat V\to\beta\hat V$, $\hat q\to\beta^{2}\hat q$, $\hat m\to\beta\hat m$, $\hat P\to\beta\hat P$. The RS channel term becomes
\begin{equation}\label{eq:replica:psi-y}
    \tilde\Psi_{y} = -\,\mathbb{E}_{\xi}\!\left[\int\dd y\,\mathcal{Z}_{0}\!\left(y,\,\tfrac{m}{\sqrt{q}}\xi,\,1-\tfrac{m^{2}}{q}\right)\mathcal{M}_{V\lossfun(y,\,\cdot\,-\advtrainingcost\sqrt[\qstar]{P})}\!\bigl(\sqrt{q}\,\xi\bigr)\right]\,,
\end{equation}
with $\mathcal{Z}_{0}(y,\omega,V)=\int(2\pi V)^{-1/2}\,e^{-(x-\omega)^{2}/(2V)}\,P_{0}(y\mid x)\,\dd x$ the conditional-channel partition function, and $\mathcal{M}_{f}$ the Moreau envelope of $f$. The RS prior term splits according to the block structure of $\featmat$ from~\Cref{ass:app:data-distribution}, yielding, for $\gamma\le1$,
\begin{equation}\label{eq:replica:psi-w-small}
    \tilde\Psi_{w}^{(\gamma\le1)} = \gamma\,\mathbb{E}_{\xi}\!\left[\mathcal{Z}_{w_\star}^{(1)}\,\log\mathcal{Z}_{w}^{(1)}\right] + (1-\gamma)\,\mathbb{E}_{\xi}\!\left[\log\mathcal{Z}_{w}^{(2)}\right]\,,
\end{equation}
and, for $\gamma>1$,
\begin{equation}\label{eq:replica:psi-w-big}
    \tilde\Psi_{w}^{(\gamma>1)} = \mathbb{E}_{\xi}\!\left[\mathcal{Z}_{w_\star}^{(3)}\,\log\mathcal{Z}_{w}^{(3)}\right]\,,
\end{equation}
where the partition functions $\mathcal{Z}_{w_\star}(\varrho,\Lambda) = \int P_{\star}(\dd w_\star)\,e^{-\Lambda w_\star^{2}/2 + \varrho w_\star}$ and $\mathcal{Z}_{w}(\varrho,\Lambda,\pi)=\int\dd w\,e^{-\lambda w^{2}/2 - \Lambda w^{2}/2 - \pi|w|^{\qstar} + \varrho w}$ are evaluated at the block-specific arguments
\begin{equation*}
\arraycolsep=3pt
\begin{array}{rl}
    \mathcal{Z}_{w_\star}^{(1)} &= \mathcal{Z}_{w_\star}\!\bigl(\tfrac{\hat m\,\xi}{\sqrt{\hat q\,(1+\gamma)}},\,\tfrac{\hat m^{2}}{\hat q\,(1+\gamma)}\bigr),\\[2pt]
    \mathcal{Z}_{w}^{(1)} &= \mathcal{Z}_{w}\!\bigl(\sqrt{\hat q\,(1+\tfrac{1}{\gamma})}\,\xi,\;\hat V\,(1+\tfrac{1}{\gamma}),\;\tfrac{\hat P}{2}\bigr),
\end{array}
\quad
\begin{array}{rl}
    \mathcal{Z}_{w}^{(2)} &= \mathcal{Z}_{w}\!\bigl(\sqrt{\hat q}\,\xi,\;\hat V,\;\tfrac{\hat P}{2}\bigr),\\[2pt]
    \mathcal{Z}_{w_\star}^{(3)} &= \mathcal{Z}_{w_\star}\!\bigl(\tfrac{\hat m\,\xi}{\sqrt{2\hat q}},\,\tfrac{\hat m^{2}}{2\hat q}\bigr),\\[2pt]
    \mathcal{Z}_{w}^{(3)} &= \mathcal{Z}_{w}\!\bigl(\sqrt{2\hat q}\,\xi,\;2\hat V,\;\tfrac{\hat P}{2}\bigr).
\end{array}
\end{equation*}
The block superscripts $(1),(2),(3)$ refer respectively to the signal block ($\gamma\le1$, dimension $d$), the kernel block ($\gamma\le1$, dimension $p-d$), and the single block ($\gamma>1$, dimension $p$) of $\featmat$.

\subsection{Saddle-point equations and agreement with~\Cref{thm:self-consistent-equations-latent-space}}

Extremizing $\Phi^{(0)}:=\lim_{r\to 0^{+}}r^{-1}\Phi^{(r)}$ in the eight RS variables $(m,V,q,P,\hat m,\hat V,\hat q,\hat P)$ gives the saddle-point system. The channel block, using the proximal-shifted residual $f_{\lossfun}(y,\omega,V,P) := V^{-1}\bigl(\mathcal{P}_{V\lossfun(y,\,\cdot\,-y\advtrainingcost\sqrt[\qstar]{P})}(\omega)-\omega\bigr)$, reads
\begin{equation}\label{eq:replica:channel-sp}
\begin{aligned}
    \hat m &= \alpha\sqrt{\gamma}\,\mathbb{E}_{\xi}\!\!\int\dd y\,\partial_\omega\mathcal{Z}_{0}\,f_{\lossfun}, &
    \hat q &= \alpha\gamma\,\mathbb{E}_{\xi}\!\!\int\dd y\,\mathcal{Z}_{0}\,f_{\lossfun}^{2},\\
    \hat V &= -\alpha\gamma\,\mathbb{E}_{\xi}\!\!\int\dd y\,\mathcal{Z}_{0}\,\partial_\omega f_{\lossfun}, &
    \hat P &= \alpha\gamma\,\advtrainingcost\,\qstar\,P^{\qstar-1}\,\mathbb{E}_{\xi}\!\!\int\dd y\,y\,\mathcal{Z}_{0}\,f_{\lossfun}.
\end{aligned}
\end{equation}
The prior block, computed from $\partial_{\hat m,\hat q,\hat V,\hat P}\tilde\Psi_{w}$ via the identities $\partial_{1}\mathcal{Z}_{w} = \mathcal{Z}_{w}f_{w}$ and $\partial_{2}\mathcal{Z}_{w} = -\tfrac{1}{2}\bigl(\partial_\varrho f_{w} + f_{w}^{2}\bigr)$, with the proximal $f_{w}(\varrho,\Lambda,\pi)=\argmin_{z}\bigl[\pi|z|^{\qstar} + \lambda z^{2} + \tfrac{\Lambda}{2}z^{2} - \varrho z\bigr]$, reproduces the prior equations~\cref{eq:latent-space-eq-prior-gamma-small,eq:latent-space-eq-prior-gamma-big} with the same case split $\gamma\le1$ versus $\gamma>1$ inherited from the block structure of~\cref{eq:app:feature-mat-definition}.

Under the change of variables $(m,\tau,V,P,\bar m,\bar\tau,\bar V,\bar P)=\bigl(m,\,q,\,V,\,P,\,\hat m/\sqrt{\gamma},\,\hat q/\gamma,\,\hat V,\,\hat P/(\alpha\gamma)\bigr)$, the channel system~\cref{eq:replica:channel-sp} coincides with~\cref{eq:latent-space-eq-channel}, and the prior system matches~\cref{eq:latent-space-eq-prior-gamma-small,eq:latent-space-eq-prior-gamma-big}. The two derivations — CGMT (rigorous, \Cref{sec:app:proof-high-d}) and replica (heuristic, this appendix) — therefore agree, providing an independent statistical-physics check of~\Cref{thm:self-consistent-equations-latent-space}.

\section{Numerical Details}\label{sec:app:numerics}

The self-consistent equations from \Cref{thm:self-consistent-equations-latent-space} are amenable to fixed-point iteration. Starting from a random initialization, we iterate through both the hat and non-hat variable equations until the maximum absolute difference between the order parameters in two successive iterations falls below a tolerance of $10^{-5}$.

To speed up convergence we use a damping scheme. Denoting by $x_i^{\rm new}$ the value of an order parameter freshly computed at iteration $i$ and by $x_{i-1}$ its value at the previous iteration, the update reads $x_i := \mu\, x_i^{\rm new} + (1-\mu)\, x_{i-1}$, where $\mu \in (0,1]$ is the damping parameter.

\revision{Once convergence is achieved for fixed $(r,\lambda)$, we jointly optimize both hyperparameters with a two-dimensional, gradient-free Nelder--Mead search over $r,\lambda\geq0$.}

For each iteration we evaluate the proximal operator numerically using SciPy's~\citep{Vir+20} Brent algorithm for univariate minimization (\texttt{scipy.optimize.minimize\_scalar}). The numerical integration is handled with SciPy's adaptive quadrature (\texttt{scipy.integrate.quad}). These routines let us evaluate the equations and the integrations to the desired accuracy.

All experiments were run on consumer-grade hardware (Mac Studio M2 Ultra, 2022), and no individual run took more than one day of CPU time.

\end{document}